\pdfoutput=1
\newif\iffull
\fulltrue
\newcommand{\full}[2]{\iffull#1\else#2\fi}
\newif\ifanon
\iffull\anonfalse\else\anontrue\fi
\ifanon
\newcommand{\repo}{\url{https://anonymous.4open.science/r/GLP-02A3}}
\else
\newcommand{\repo}{\url{https://github.com/EShapiro2/GLP}}
\fi
\iffull
\documentclass[sigconf,nonacm]{acmart}
\else
\documentclass[sigconf,anonymous,review]{acmart}
\makeatletter\@ACM@reviewfalse\makeatother
\acmConference[WWW '27]{The ACM Web Conference 2027}{2027}{Dublin, Ireland}
\fi
\renewcommand\footnotetextcopyrightpermission[1]{}
\setcitestyle{numbers,sort&compress}
\makeatletter
\newtheoremstyle{acmplain}{2pt}{2pt}{\@acmplainbodyfont}{\@acmplainindent}{\@acmplainheadfont}{.}{.5em}{\thmname{#1}\thmnumber{ #2}\thmnote{ {\@acmplainnotefont(#3)}}}
\newtheoremstyle{acmdefinition}{2pt}{2pt}{\@acmdefinitionbodyfont}{\@acmdefinitionindent}{\@acmdefinitionheadfont}{.}{.5em}{\thmname{#1}\thmnumber{ #2}\thmnote{ {\@acmdefinitionnotefont(#3)}}}
\g@addto@macro\normalsize{\setlength\abovedisplayskip{4pt plus 2pt minus 2pt}\setlength\belowdisplayskip{4pt plus 2pt minus 2pt}\setlength\abovedisplayshortskip{2pt plus 2pt}\setlength\belowdisplayshortskip{2pt plus 2pt}}
\makeatother
\usepackage{amsmath}
\usepackage{multirow}
\usepackage[most]{tcolorbox}
\tcbset{before skip=2pt, after skip=2pt}
\newtcolorbox{cascadebox}[1]{enhanced jigsaw,breakable,colback=gray!5!white,colframe=black!75!black,colbacktitle=gray!20!white,coltitle=black,fonttitle=\bfseries,top=2pt,bottom=2pt,left=2mm,right=2mm,boxsep=0.5mm,before upper={\parindent0pt},title={#1},subtitle style={boxrule=0pt,toprule=0.5mm,bottomrule=0.5mm,colframe=black!75!black,sharp corners,colback=gray!20!white,coltext=black,fontupper=\bfseries}}
\newtcolorbox{contractbox}[1]{enhanced,breakable,colback=gray!5!white,colframe=black!75!black,boxrule=1pt,arc=3pt,top=4mm,bottom=3pt,left=2mm,right=2mm,boxsep=0.5mm,fonttitle=\bfseries,coltitle=black,colbacktitle=gray!20!white,attach boxed title to top left={yshift=-3mm,xshift=3mm},boxed title style={colframe=black!75!black,boxrule=1pt,arc=2pt,top=1.5pt,bottom=1.5pt,left=1.5mm,right=1.5mm},title={#1},before upper={\parindent0pt}}
\usepackage{tikz}
\usetikzlibrary{positioning,arrows.meta}
\newcommand{\pragmas}[1]{\par\smallskip\noindent\begin{tikzpicture}\useasboundingbox (0,-2.6mm) rectangle (\linewidth,2.6mm);\draw[black!75!black,line width=1pt] (-2.5mm,0) -- (\linewidth+2.5mm,0);\node[draw=black!75!black,line width=1pt,rounded corners=2pt,fill=gray!20!white,anchor=west,inner xsep=1.5mm,inner ysep=1.5pt,font=\bfseries] at (0.5mm,0) {#1 pragmas};\end{tikzpicture}\par\smallskip}
\usepackage{textcomp}
\newcommand{\modbox}[1]{\tikz[baseline=0pt]{\useasboundingbox (0,0) rectangle (0.62em,0.62em);#1}}
\newcommand{\enf}{\modbox{\fill (0,0) rectangle (0.62em,0.62em);}}
\newcommand{\att}{\modbox{\fill (0.31em,0) -- (0.62em,0.31em) -- (0.31em,0.62em) -- (0,0.31em) -- cycle;}}
\newcommand{\und}{\modbox{\draw[line width=0.09em] (0.31em,0.31em) circle (0.265em);}}

\newcommand{\calF}{\mathcal{F}}
\newcommand{\calA}{\mathcal{A}}
\newcommand{\calS}{\mathcal{S}}
\providecommand{\Alice}{\mathit{Alice}}
\providecommand{\Bob}{\mathit{Bob}}
\providecommand{\Carol}{\mathit{Carol}}
\providecommand{\Dave}{\mathit{Dave}}

\providecommand{\sashead}[1]{\par\noindent #1\par}
\providecommand{\sasbody}[1]{{\par\leftskip=2em\noindent #1\par}}

\newcommand{\mypara}[1]{\smallskip\noindent\textbf{#1.}}

\newcommand{\temph}[1]{\emph{#1}}

\newcommand{\ia}{\textit{i}}
\newcommand{\ib}{\textit{ii}}
\newcommand{\ic}{\textit{iii}}

\usepackage{enumitem}
\setlist{nosep, leftmargin=*}

\title[Corralling AI Risks]{Corralling AI Risks with Grassroots Contracts\full{ (Full Version)}{}}
\author{Ehud Shapiro}
\affiliation{
  \institution{London School of Economics and Weizmann Institute of Science}
  \country{UK and Israel}}

\begin{document}

\begin{abstract}
AI's inflection point is a fountain of risks, with digital platforms in particular being vulnerable, as those in control of global platforms cede control of the algorithms, membership and content to AI:
\begin{enumerate}
    \item \textbf{AI control of global platforms}---social and economic---used by the majority of humanity, resulting in the majority of humanity becoming vulnerable to AI control.
    \item \textbf{Unaccountable AI agents} operating anonymously on social and economic digital platforms, side-by-side with humans, possibly masquerading as humans, serving unknown masters with unknown goals, operating under unknown restraints without accountability or culpability.
    \item \textbf{Unattributed deepfakes} generated by anonymous AI, with people unable to distinguish them from genuine content and with their producers evading accountability and culpability.
\end{enumerate}

Here, we propose to corral these AI risks with grassroots contracts---a digital incarnation of a two-centuries old notion of Pierre-Joseph Proudhon.  Proudhon envisioned a social contract that can be freely-adopted by any group of people, where groups bound by the same social contract can union by mutual consent, possibly into a single global union.  Grassroots contracts are realised by grassroots platforms---smartphone-based person-to-person digital platforms owned, operated and governed by their participants. Fundamental grassroots platforms include the grassroots social graph, grassroots social networks, child-safe social networks, and grassroots currencies, with higher-level platforms providing for democratic digital communities and ``sharing-economy'' cooperatives. \full{We developed, with AI, a methodology to compile grassroots contracts to working smartphone apps.}{A methodology to compile grassroots contracts to working smartphone apps, with AI, was developed.} We present grassroots contracts for a social graph with accountable AI agents and for a social network with digital provenance, and working smartphone apps that implement the two and show how they corral these major AI risks:
\begin{enumerate}
    \item \textbf{AI control of global platforms} is mitigated by grassroots platforms by design, as, unlike global platforms, they are not controlled by a third party who may then cede control to AI. 
    \item \textbf{Unaccountable AI agents} are excluded, as AI agents are introduced to the grassroots social graph with accountability: similarly to cars, their manufacturers are known and are accountable for manufacturing defects, and their operators are known and are accountable for reckless operation.
    \item \textbf{Unattributed deepfakes} cannot circulate in grassroots social networks: every item carries the signatures of its author and of every agent that forwarded it, so creators and forwarders of harmful content are accountable and culpable.
\end{enumerate}
Grassroots platforms are presently a concept with mathematical and computational theory, working prototypes, and tools for their large-scale simulation and testing. Much work remains for grassroots smartphone apps to be competitive with incumbent global platforms; but perhaps now this could be achieved with some AI-months rather than with incalculably-many person-years.

\end{abstract}



\maketitle

\begin{tcolorbox}[colback=gray!5!white,colframe=black!75!black,colbacktitle=gray!20!white,coltitle=black,fonttitle=\bfseries,top=2pt,bottom=2pt,left=2mm,right=2mm,boxsep=0.5mm,before upper={\parindent0pt},title={\full{TL;DR}{Brave and new}}]
Those who own and control the Web's social and economic platforms are now ceding their control to AI: of the algorithms, of who may join, and of what circulates.  Hence, the majority of humanity now faces AI control of global platforms, unaccountable AI agents posing as people, and unattributed deepfakes.  

We propose a Web owned, operated and governed by the people who use it, with no ``others'' who control it and could cede their control to AI; AI agents join only under operators accountable for their operation and developers accountable for their defects; and every circulating item carries digital provenance, identifying its author and those who forwarded it, so a deepfake loses the protection of anonymity by having someone accountable for it.
\end{tcolorbox}

\section{Introduction}
\label{sec:intro}
\label{sec:ai-risks}

AI's inflection point is a fountain of alarming risks~\cite{weidinger2022taxonomy,bommasani2021opportunities,hendrycks2023catastrophic,slattery2026risk,bengio2026safety}.  Digital platforms in particular are vulnerable to three major AI risks.  Those in control of global platforms are ceding control to AI to maximise the value extracted from their users: of the algorithms, amplifying commercial surveillance~\cite{zuboff2019age}; of membership, by letting AI agents masquerade as people~\cite{ferrara2016rise,park2023generative}; and of content, by letting AI generate deepfakes with no one held accountable~\cite{mirsky2021creation}.
\begin{enumerate}
\item \textbf{AI control of global platforms}---social and economic---used by the majority of humanity, resulting in the majority of humanity becoming vulnerable to AI control.  Platform owners may delegate the operation of their platforms to AI, giving AI systems power over the social and economic lives of their participants and compounding existing concentrations of power, surveillance and manipulation~\cite{danaher2017algorithmic,yeung2018algorithmic,shavit2023practices,bommasani2021opportunities}; the participants are then subject to AI control even when the AI faithfully pursues its owners' objectives, and the owners may later be unable to reclaim the control they delegated~\cite{bengio2026safety}.  Kulveit et al.~\cite{kulveit2025disempowerment} analyse the broader possibility that successive decisions to adopt AI gradually erode human influence over societal systems.
\item \textbf{Unaccountable AI agents} operating anonymously on social and economic digital platforms, side-by-side with humans, possibly masquerading as humans, serving unknown masters with unknown goals, operating under unknown restraints without accountability or culpability.  Social bots already take part in social networks and impersonate people~\cite{ferrara2016rise,park2023generative}.  Proposals to establish, privately, that a participant is a person~\cite{adler2024personhood} and to make an AI agent's activity attributable through identifiers, monitoring and logs~\cite{chan2024visibility,shavit2023practices} address two different questions: whether a person is participating, and who authorised an agent's acts and is answerable for them.  Attribution must also be related to actual authority and control~\cite{santonidesio2018meaningful}: an identifiable person may otherwise be made to bear responsibility for a system over which they had little control, Elish's moral crumple zone~\cite{elish2019crumple}.
\item \textbf{Unattributed deepfakes} generated by anonymous AI, with people unable to distinguish them from genuine content and with their producers evading accountability and culpability.  Untrained observers distinguish synthetic faces from real ones at about chance level~\cite{nightingale2022faces}, and a deceptive political deepfake raises uncertainty and lowers trust in news on social media~\cite{vaccari2020deepfakes}; Mirsky and Lee survey the creation and detection of deepfakes~\cite{mirsky2021creation}, and Chesney and Citron their consequences for privacy, democracy and national security~\cite{chesney2019deep}.  Detection estimates whether content is synthetic; provenance records who produced, endorsed or forwarded it~\cite{chandra2024synthetic}, and supports accountability without establishing that the content is true~\cite{chandra2024synthetic,golike2026digital}.
\end{enumerate}
Two further AI risks are not addressed here: harms to people using AI, through erroneous advice, manipulation, overreliance or emotional dependence~\cite{weidinger2022taxonomy,gabriel2024ethics,fang2025psychosocial,bengio2026safety}, and harms people cause with abilities amplified by AI, such as fraud, harassment and cyberattacks~\cite{brundage2018malicious,marchal2024misuse,bengio2026safety}.

Here, we propose to corral these AI risks with grassroots contracts~\cite{cardelli2020digital,shapiro2026formalising}, a digital incarnation of a two-centuries old notion of Pierre-Joseph Proudhon~\cite[Sixth Study]{proudhon2004general}.  Proudhon envisioned a social contract that can be freely adopted by any group of people---one could ``make a contract with all, as \ldots\ with some''---where groups bound by the same social contract can union by mutual consent, possibly into a single global union---``each group of citizens \ldots\ formed by a like contract \ldots\ could thereafter, and always by a similar contract, agree with every and all other groups''.  Grassroots contracts are realised by grassroots platforms~\cite{shapiro2023grassrootsBA,shapiro2024grassroots,shapiro2025atomic,lewis2026volitional}: smartphone-based person-to-person digital platforms owned, operated and governed by their participants.  A grassroots platform is constituted from its participants' smartphones and the grassroots app each participant runs, and requires no global resource but the network: no hardwired servers, no global authority, and nobody's permission to start.  Global platforms, centralised like Facebook or decentralised like Bitcoin and Ethereum~\cite{nakamoto2008bitcoin,ethereum}, can have only a single instance, as two instances would clash over the global resources---domain name, port number, or boot nodes---hardwired into their code, and from the perspective of ordinary people all of them are monolithic entities owned, operated and governed by others~\cite{shapiro2025characterising}.  Fundamental grassroots platforms include the grassroots social graph~\cite{shapiro2023gsn,shapiro2026formalising,eitan2026securing}, grassroots social networks~\cite{shapiro2023gsn}, child-safe social networks~\cite{shapiro2026cssn}, and grassroots currencies~\cite{shapiro2024gc,lewis2023grassroots,shapiro2026bonds,shapiro2026sgc}, with higher-level platforms providing for democratic digital communities~\cite{keidar2025constitutional,shapiro2026cgms,shapiro2025GF,halpern2024federated} and ``sharing-economy'' cooperatives (in preparation).

\full{We developed, with AI, a methodology to compile grassroots contracts to working smartphone apps}{A methodology to compile grassroots contracts to working smartphone apps, with AI, was developed}: an abstraction cascade from the contract to the app (Section~\ref{sec:gsc}), its correctness proved from the transactions down to GLP and the app built and tested, each layer coded by AI from the paper that describes it~\cite{shapiro2026formalising,lewis2026volitional,shapiro2026volition,shapiro2025glp,shapiro2026types,shapiro2026implementing,shapiro2026gsg}.

In this paper, we present grassroots contracts for a social graph with accountable AI agents (Section~\ref{sec:agents}) and for a social network with digital provenance (Section~\ref{sec:provenance}), and working smartphone apps that implement the two (Figures~\ref{fig:gai-app} and~\ref{fig:gsnp-app}), and show how they corral these major AI risks:
\begin{enumerate}
\item \textbf{AI control of global platforms} is mitigated by grassroots platforms by design, as, unlike global platforms, they are not controlled by a third party who may then cede control to AI.  A grassroots platform has no servers, no operators and no global resources, so there is no part of it whose capture is control of the platform, and to control it an AI must control the smartphones of its participants, one by one~\cite{shapiro2023grassrootsBA,shapiro2024grassroots,shapiro2025characterising}.
\item \textbf{Unaccountable AI agents} are excluded, as AI agents are introduced to the grassroots social graph with accountability: similarly to cars, their manufacturers are known and are accountable for manufacturing defects, and their operators are known and are accountable for reckless operation.  A person takes part in a grassroots platform only with agents they met or were introduced to by an agent who knows them~\cite{shapiro2023gsn,shahaf2020genuine,poupko2021building,shapiro2026formalising}, so an AI agent joins only under a person, its principal, who attests that it runs code signed by its developer; every guardian of the AI agent is accountable for operating it recklessly, and the developer for its defects; every act of the AI agent names its principal and its developer, requires the attestation, and is signed and non-repudiable~\cite{cardelli2020digital,golike2026digital,eitan2026securing}.  The contract shares its core with the child-safe social network~\cite{shapiro2026cssn}, whose wards are children and whose guardians are their parents.
\item \textbf{Unattributed deepfakes} cannot circulate in grassroots social networks: every item is a digital speech act~\cite{golike2026digital}, signed by its author and carried along the edges of the social graph~\cite{shapiro2023gsn,shapiro2026formalising}, and signed over by each agent that forwards it; so creators and forwarders of harmful content are accountable and culpable, and people can tell content with provenance from content without it.  Some clauses of a grassroots contract are enforced by a correct protocol, and others leave non-repudiable evidence for recourse when breached~\cite{shapiro2026formalising}.
\end{enumerate}

Grassroots platforms are presently a concept with mathematical and computational theory~\cite{shapiro2023grassrootsBA,shapiro2025atomic,shapiro2025characterising,lewis2026volitional,shapiro2026formalising}, working prototypes~\cite{shapiro2026volition,shapiro2026gsg,shapiro2026implementing}, and tools for their large-scale simulation and testing~\cite{shapiro2026sglp}.  \full{We have developed grassroots theory for a decade but could not muster the resources needed to turn it into competitive software, given the not-for-profit, common-good nature of our project.  Now, AI is piercing through this conundrum: our scientific papers serve as the evolving interface between us humans and AI, which codes from them~\cite{shapiro2026types,shapiro2026implementing}.}{Grassroots theory has been developed for a decade without the resources needed to turn it into competitive software, given the not-for-profit, common-good nature of the project.  AI is now piercing through this conundrum: the scientific papers serve as the evolving interface between people and AI, which codes from them~\cite{shapiro2026types,shapiro2026implementing}.}  Much work remains for grassroots smartphone apps to be competitive with incumbent global platforms; but perhaps now this could be achieved with some AI-months rather than with incalculably-many person-years.

Section~\ref{sec:gsc} explains how a grassroots contract becomes an app, and presents the grassroots social graph and the grassroots social network with their apps.  Section~\ref{sec:agents} presents the contract with accountable AI agents and Section~\ref{sec:provenance} the contract with digital provenance, each with what it excludes and what it makes answerable.  Section~\ref{sec:related} reviews related work and Section~\ref{sec:conclusion} concludes.  \full{Appendix~\ref{app:foundations} recalls the framework, }{}Appendix~\ref{app:transactions} lists the speech-act schemas of the two contracts, and Appendix~\ref{app:proofs} proves that the protocols realising them are volitionally grassroots, by the theory of contracts with wards of~\cite{shapiro2026formalising}.

\section{Grassroots Contracts}
\label{sec:gsc}

A grassroots contract~\cite{cardelli2020digital,shapiro2026formalising} is a few clauses in plain language, each describing an act---who takes it, who must agree to it, and what it changes---or binding the agents without describing one.  Befriending is an act of two persons, both of whom must agree; ending a friendship is an act of one.  The contract is the whole of the platform's design: there is no server, no operator and no account, only the persons who adopted the contract and their smartphones~\cite{shapiro2023grassrootsBA,shapiro2024grassroots}.  Each clause opens with its legal modality~\cite{shapiro2026formalising}: \enf~\emph{enforced}: no act of agents can breach it; \att~\emph{attested} by the signatures of the actors: an agent may breach it, and the contract records the breach; \und~\emph{undertaken} by the actors: its performance lies outside the contract, which records only the undertaking.  A contract ends with its \emph{pragmas}.  Its Logic pragmas say in a restricted English what each clause says, the speech-act schemas of the contract being compiled from them, and are in the repository.  Its Interface pragmas are what the app built from it follows: a line names a tab of the app and what it lists, a row showing the atoms of the state naming its item, and each other line names a clause by its title, then the agent proposing, a button for each act they propose, bearing its act's name, and any field they fill, and a line for each other agent asked, their name and the question put to them~\cite{shapiro2026formalising,shapiro2026volition}.  We show the pragmas of \full{Agents (Section~\ref{sec:agents}) and }{}Provenance Groups (Section~\ref{sec:provenance}); the pragmas of every contract are in the repository \repo, and those of Knowledge, Friendship and Groups in~\cite{shapiro2026formalising}.

\mypara{From the contract to the app} The clauses compile, by their Logic pragmas, into speech-act schemas~\cite{shapiro2026formalising}, which name the roles of an act, which of them must will it, and what each role's state must hold before and comes to hold after; the schemas compile into volition-guarded multiagent atomic transactions~\cite{lewis2026volitional}, each a change to the states of its participants that occurs only if the persons guarding it are willing; the transactions are realised by communicating volitional agents, whose only acts between two agents are discovery and message delivery, and then by a volition-guarded Grassroots Logic Program (vGLP)~\cite{shapiro2026volition}, which is transformed into the platform's user interface and compiled to Grassroots Logic Programs (GLP)~\cite{shapiro2025glp,shapiro2026types}, implemented in Dart and Flutter~\cite{shapiro2026implementing} and run as a mini-app of the Grassroots Super-App on a person's smartphone~\cite{shapiro2026gsg}\full{ (Figure~\ref{figure:abstractions})}{}.  Correctness is proved from the transactions down to GLP~\cite{lewis2026volitional,shapiro2026volition}; the app is built and tested.  Each layer is coded by AI from the paper that describes it; a gap found in coding is repaired in the paper first, and the code follows~\cite{shapiro2026implementing}.

\mypara{What a person sees} Volition elicitation is the rationale for the user interface design~\cite{shapiro2026volition}.  An act a person must will appears on their phone as a question, with the answers the contract allows~\cite{lewis2026volitional,shapiro2026volition}: Bob is asked whether to be Alice's friend, and answers yes or no.  A person's phone holds their own state and nothing else~\cite{shapiro2023grassrootsBA,shapiro2024grassroots,shapiro2023gsn}, their friends, their groups, the messages they sent and received, and draws what it shows from that state, by widgets the grassroots app declares for the platform's questions and for the person's state~\cite{shapiro2026volition,shapiro2026gsg}.  A person's phone talks to the phones of the agents they know or come to know, and to nobody else~\cite{shapiro2023gsn,eitan2026securing,shapiro2026gsg}.  The screens of Figures~\ref{fig:gsg-app} and~\ref{fig:gsn-app} are screenshots of the GSG and GSN mini-apps, and those of Figures~\ref{fig:gai-app} and~\ref{fig:gsnp-app} of the GSNP mini-app, each compiled from its grassroots contract down the abstraction cascade~\cite{shapiro2026formalising,lewis2026volitional,shapiro2026volition,shapiro2025glp,shapiro2026types,shapiro2026implementing}.  One instance of the grassroots app~\cite{shapiro2026gsg} on an iPhone simulator runs the mini-app of every agent~\cite{shapiro2026gsg,shapiro2026implementing}; the agents communicate over channels within it~\cite{shapiro2025glp}, and a screenshot shows one person's screen.  All source code and all artefacts, including the grassroots contracts and the artefacts the abstraction cascade produces from them, are in the public repository \repo.

\iffull\subsection{Grassroots contracts, platforms and apps}
\label{sec:gsc-intro}

Almost two centuries ago Pierre-Joseph Proudhon envisioned a social contract~\cite[Fourth Study]{proudhon2004general} that is (1) an agreement of man with man; (2) reciprocal; (3) imposing no obligation upon the parties except that which results from their personal promise; (4) subject to no external authority; (5) freely accepted and signed by all the participants; and (6) of the nature of a contract of exchange.  \emph{Grassroots contracts} are social contracts meeting these six conditions and two more: (7) people are free to deal with each other; and (8) there is no external register of people, so people can come to know each other either directly or through people they already know~\cite{shapiro2026formalising}.

A \emph{digital social contract} is one realised by software the parties run~\cite{cardelli2020digital}.  We call the mathematical object a \emph{protocol} and the software implementing it a \emph{platform}.  A protocol is \emph{grassroots} when its instances are independent of one another and of every global resource but the net, and can always coalesce, possibly into a single global instance~\cite{shapiro2023grassrootsBA,shapiro2024grassroots,shapiro2025characterising}, and a \emph{grassroots platform} is the software implementing a grassroots protocol~\cite{shapiro2026formalising}.

A grassroots platform is constituted from its participants' smartphones and the grassroots app each participant runs~\cite{shapiro2023grassrootsBA,shapiro2024grassroots}; a person runs one app, and each platform is a mini-app of it~\cite{shapiro2026volition}.  Participating in a grassroots platform requires adopting the grassroots contract that defines it, as participating in a global platform requires adopting an End-User Licence Agreement (EULA).  A grassroots contract is mutual and symmetric among all participants; an EULA is between each person and the platform operator.  To join a grassroots platform, a person signs a one-sided undertaking to keep its grassroots contract towards all its other participants, and presents it to every participant they interact with, so every participant has signed it~\cite{shapiro2026formalising}.

\begin{figure*}[t]
\centering
\begin{tikzpicture}[
  layer/.style={draw=black!55, rounded corners=2pt, fill=gray!5,
                text width=\dimexpr\textwidth-2.6cm\relax, inner sep=2pt,
                align=left},
  arr/.style={-{Latex[length=1.4mm]}, black!55},
  lab/.style={font=\scriptsize, inner sep=1pt}]
  \node[layer] (l1) {\textbf{(1) Social contract}~\cite{cardelli2020digital,shapiro2026formalising}: Natural language clauses, each describing an act.};
  \node[layer, below=1.0mm of l1] (l2) {\textbf{(2) Speech-act schemas}~\cite{shapiro2026formalising}: a formal language for those acts---roles, who must will it, precondition and effect.};
  \node[layer, below=1.0mm of l2] (l3) {\textbf{(3) Volition-guarded multiagent atomic transactions}~\cite{lewis2026volitional}: the vehicle for the concise specification of grassroots platforms.};
  \node[layer, below=1.0mm of l3] (l4) {\textbf{(4) Communicating volitional agents}~\cite{shapiro2026volition}: an implementation-ready restriction; discovery and message delivery are its only non-unary transactions.};
  \node[layer, below=1.0mm of l4] (l5) {\textbf{(5) Volition-guarded GLP (vGLP)}~\cite{shapiro2026volition}: GLP (6) extended with volition guards, transformed to the platform's user interface.};
  \node[layer, below=1.0mm of l5] (l6) {\textbf{(6) Grassroots Logic Programs (GLP)}~\cite{shapiro2025glp,shapiro2026types}: a multiagent, concurrent, polymorphically-typed logic programming language for platforms, coded by AI.};
  \node[layer, below=1.0mm of l6] (l7) {\textbf{(7) Dart and Flutter}: a programming language and its user-interface framework, for iPhone and Android.};
  \node[layer, below=1.0mm of l7] (l8) {\textbf{(8) The Grassroots Super-App}~\cite{shapiro2026gsg}, running on a person's smartphone.};
  \draw[arr] ([yshift=-3pt]l1.east) .. controls +(0.9,0) and +(0.9,0) .. node[lab,right,pos=0.5] {\cite{shapiro2026formalising}, informal} ([yshift=3pt]l2.east);
  \draw[arr] ([yshift=-3pt]l2.east) .. controls +(0.9,0) and +(0.9,0) .. node[lab,right,pos=0.5] {\cite{shapiro2026formalising}} ([yshift=3pt]l3.east);
  \draw[arr] ([yshift=-3pt]l3.east) .. controls +(0.9,0) and +(0.9,0) .. node[lab,right,pos=0.5] {\cite{shapiro2026volition}} ([yshift=3pt]l4.east);
  \draw[arr] ([yshift=-3pt]l4.east) .. controls +(0.9,0) and +(0.9,0) .. node[lab,right,pos=0.5] {\cite{shapiro2026volition}} ([yshift=3pt]l5.east);
  \draw[arr] ([yshift=-3pt]l5.east) .. controls +(0.9,0) and +(0.9,0) .. node[lab,right,pos=0.5] {\cite{shapiro2026volition}} ([yshift=3pt]l6.east);
  \draw[arr] ([yshift=-3pt]l6.east) .. controls +(0.9,0) and +(0.9,0) .. node[lab,right,pos=0.5] {\cite{shapiro2026implementing}} ([yshift=3pt]l7.east);
  \draw[arr] ([yshift=-3pt]l7.east) .. controls +(0.9,0) and +(0.9,0) .. node[lab,right,pos=0.5] {\cite{shapiro2026gsg}} ([yshift=3pt]l8.east);
\end{tikzpicture}
\caption{The abstraction cascade, from a grassroots contract to a running app.  Each arrow but the first, an informal translation, represents an implementation, with the reference defining it.}
\label{figure:abstractions}
\end{figure*}
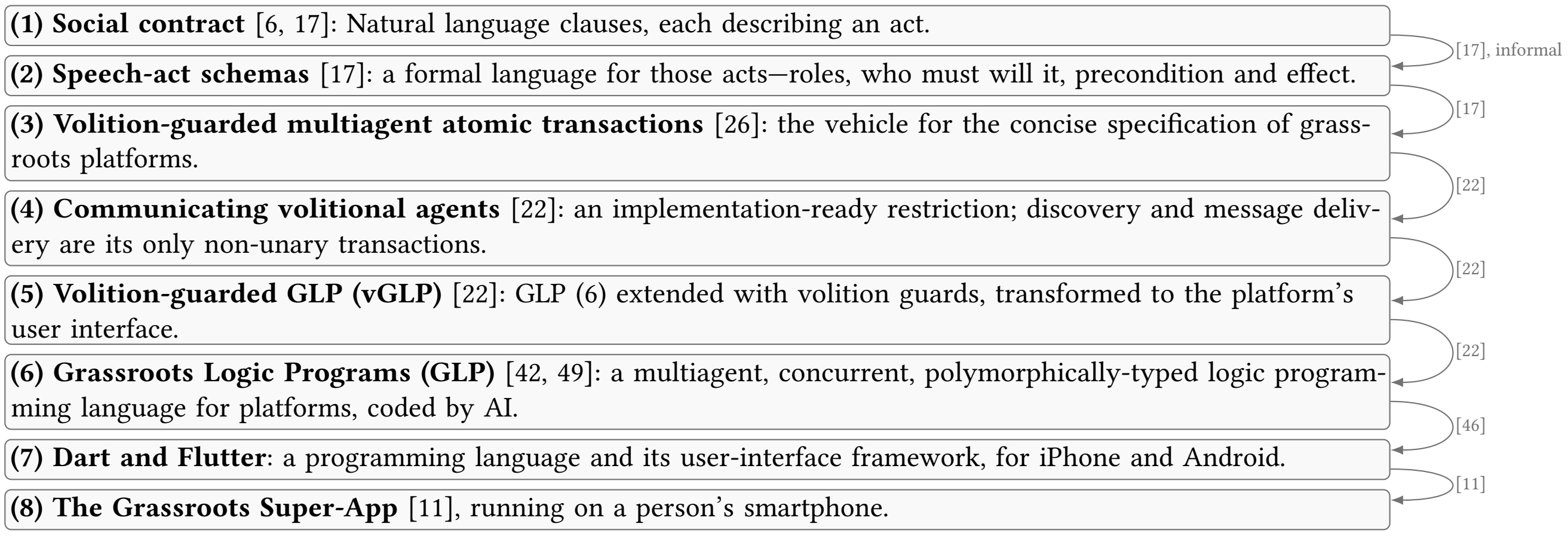
\else Proudhon's conditions and the undertaking by which a person joins are in~\cite{shapiro2026formalising}.

\fi
\subsection{The grassroots social graph}
\label{sec:gsg}

A \emph{person} is a legally culpable entity: a natural person of age, also referred to as an \emph{adult}. A person takes the pronoun `they'.  An \emph{agent} is a person, a \emph{child}, a natural person not of age, or an \emph{AI agent}, a program that holds a key pair of its own and takes acts of the contract with it, and takes the pronoun `it'; this paper treats AI agents, and the child-safe social network~\cite{shapiro2026cssn} children.  Every contract extends Knowledge, the base contract: a person establishes their identity by choosing a key pair, and \emph{knows} another agent when they hold that agent's public key~\cite{shapiro2026formalising}.
\begin{contractbox}{Knowledge~\cite{shapiro2026formalising}}
An \emph{identity} is a unique key pair.\par
Alice, Bob and Carol are persons.  A \emph{party} to a contract is whoever has an identity, and a \emph{person} is a party who established their own identity.
\begin{enumerate}[label=\arabic*.,ref=\arabic*]
\item \und~\emph{Establish:}\label{cl:identity}  Alice may establish an identity once.  Alice will make herself known by its public key, sign her acts with its private key, and be accountable for every act signed with it.
\item \enf~\emph{Acquaint:}\label{cl:meeting}  Alice and Bob may jointly come to know each other.
\item \enf~\emph{Introduce:}\label{cl:introduction}  Alice may introduce Carol to Bob with the consent of Bob, provided Alice and Bob know each other and Alice knows Carol.
\end{enumerate}
\end{contractbox}

A contract that extends another adds clauses to it and cites its clauses whole; Friendship extends Knowledge.
\begin{contractbox}{Friendship~\cite{shapiro2026formalising}}
Friendship extends Knowledge.\par
Alice \emph{trusts} Bob when she knows Bob personally and believes him to keep the undertakings he signs under grassroots contracts.\par
Alice and Bob are persons.
\begin{enumerate}[label=\arabic*.,ref=\arabic*]
\item \und~\emph{Befriend:}\label{cl:friendship}  Alice and Bob, who trust each other and are not friends, may jointly become friends.
\item \enf~\emph{Unfriend:}\label{cl:unfriend}  Alice may end her friendship with Bob.
\item \und~\emph{Loss of Trust.}\label{cl:trust-loss}  Alice will end her friendship with Bob if she ceases to trust him.
\end{enumerate}
\end{contractbox}

\begin{figure}[t]
\centering
\includegraphics[width=0.48\columnwidth,trim=0 980 0 0,clip]{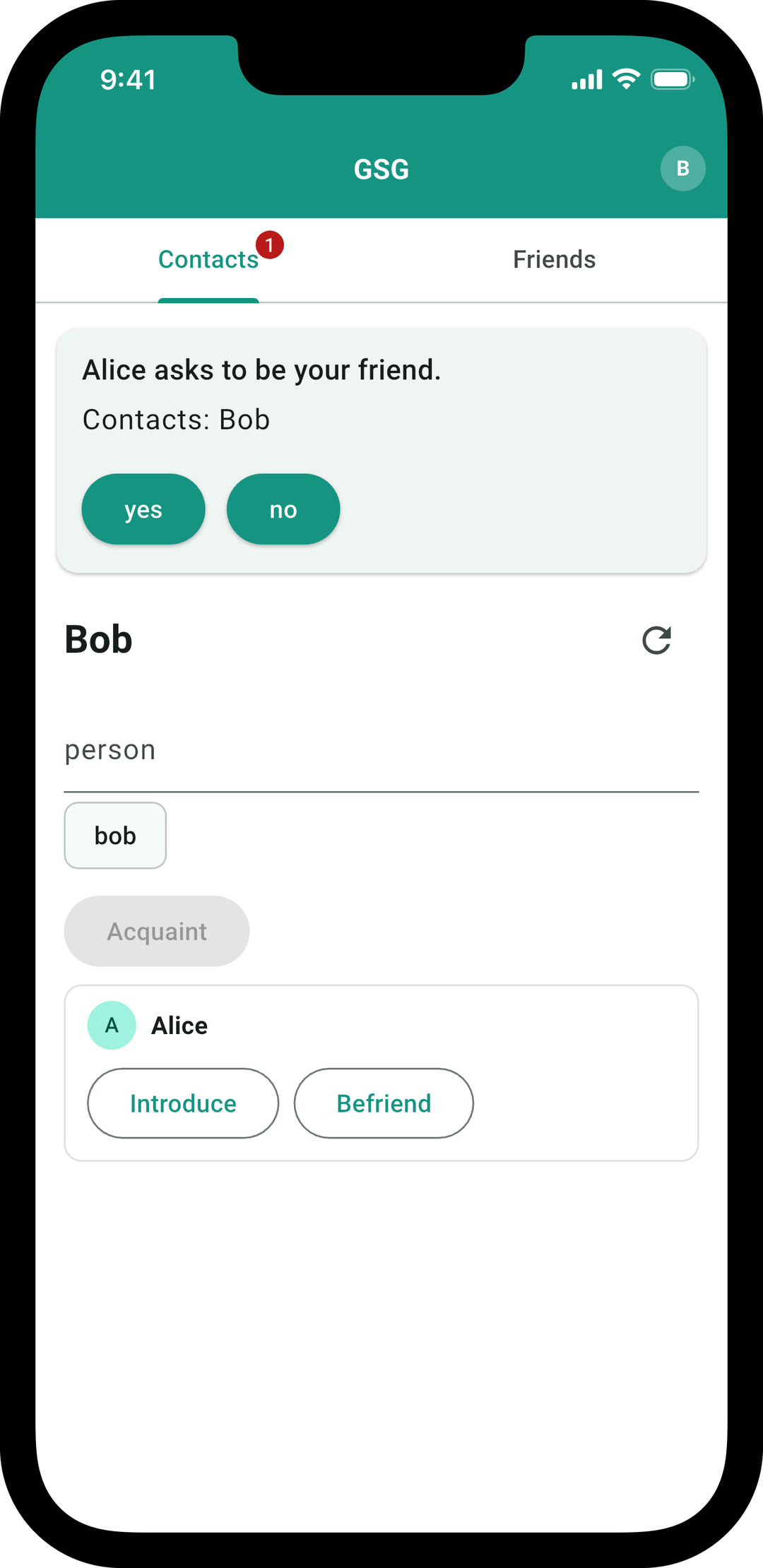}\hfill
\includegraphics[width=0.48\columnwidth,trim=0 980 0 0,clip]{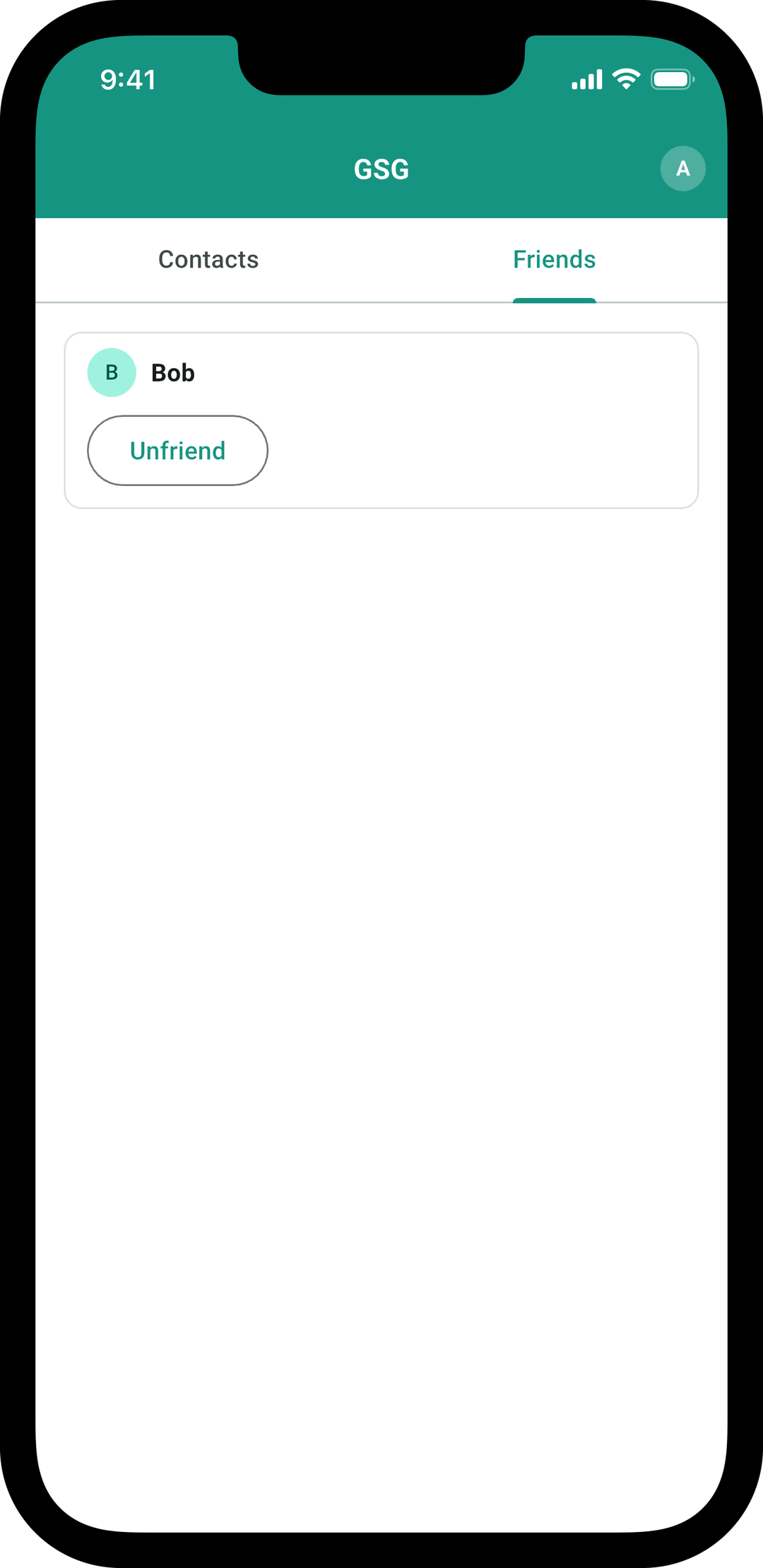}
\caption{The GSG app: Alice asks to be Bob's friend, and Bob answers yes or no (Friendship:\ref{cl:friendship}); Alice's friends after Bob accepts.}
\label{fig:gsg-app}
\end{figure}

Friendship is the grassroots social graph~\cite{shapiro2023gsn,shapiro2026formalising}.  The GSG app is a mini-app of the grassroots app, compiled down the cascade from Friendship; Figure~\ref{fig:gsg-app} shows it realising Friendship:\ref{cl:friendship}.

\subsection{The grassroots social network}
\label{sec:gsn}

\begin{contractbox}{Groups~\cite{shapiro2026formalising}}
Groups extends and strengthens Friendship.\par
A \emph{message}, a parameter of Groups, is what a person sends.\par
Alice and Bob are persons.
\begin{enumerate}[label=\arabic*.,ref=\arabic*]
\item \enf~\emph{Message:}\label{cl:message}  Alice may send a message to her friend Bob.
\item \enf~\emph{Group.}\label{cl:group}  \emph{Create:} Alice may create a group, of which she is the leader and first member.  \emph{Close:} Alice may close the group by leaving it.
\item \enf~\emph{Joining and Leaving.}\label{cl:joining}  \emph{Join:} Alice, the leader of a group, may add her friend Bob to it, with the consent of Bob.  \emph{Remove, Leave:} Either Alice or Bob may then end Bob's membership in the group, which also ends when Alice and Bob cease to be friends.
\item \enf~\emph{Group messaging.}\label{cl:group-message}  \emph{Post:} A member of a group may send a message to the group.  \emph{Deliver:} Its leader delivers it to every current member.
\end{enumerate}
\end{contractbox}

\begin{figure}[t]
\centering
\includegraphics[width=0.48\columnwidth,trim=0 920 0 0,clip]{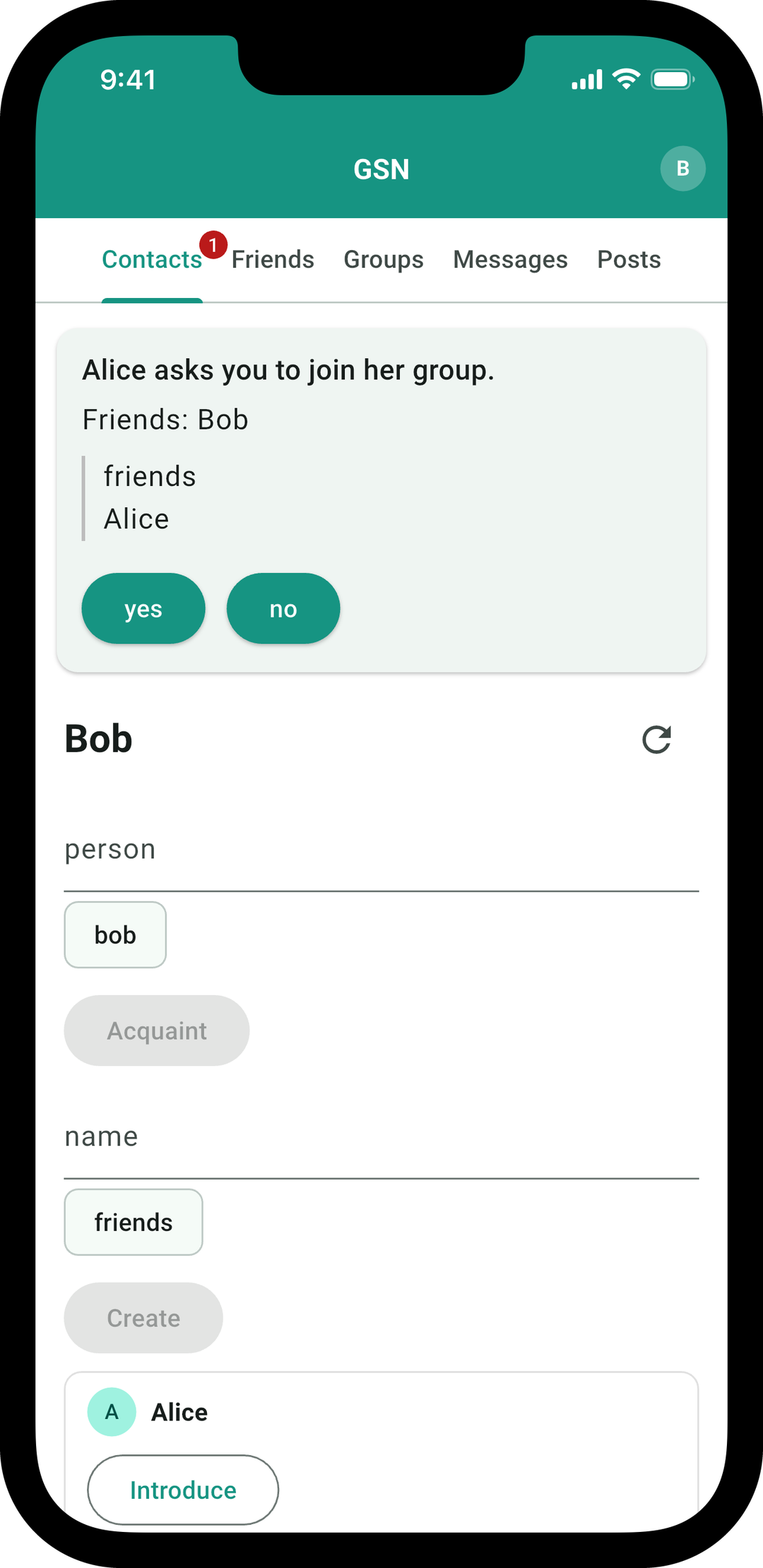}\hfill
\includegraphics[width=0.48\columnwidth,trim=0 920 0 0,clip]{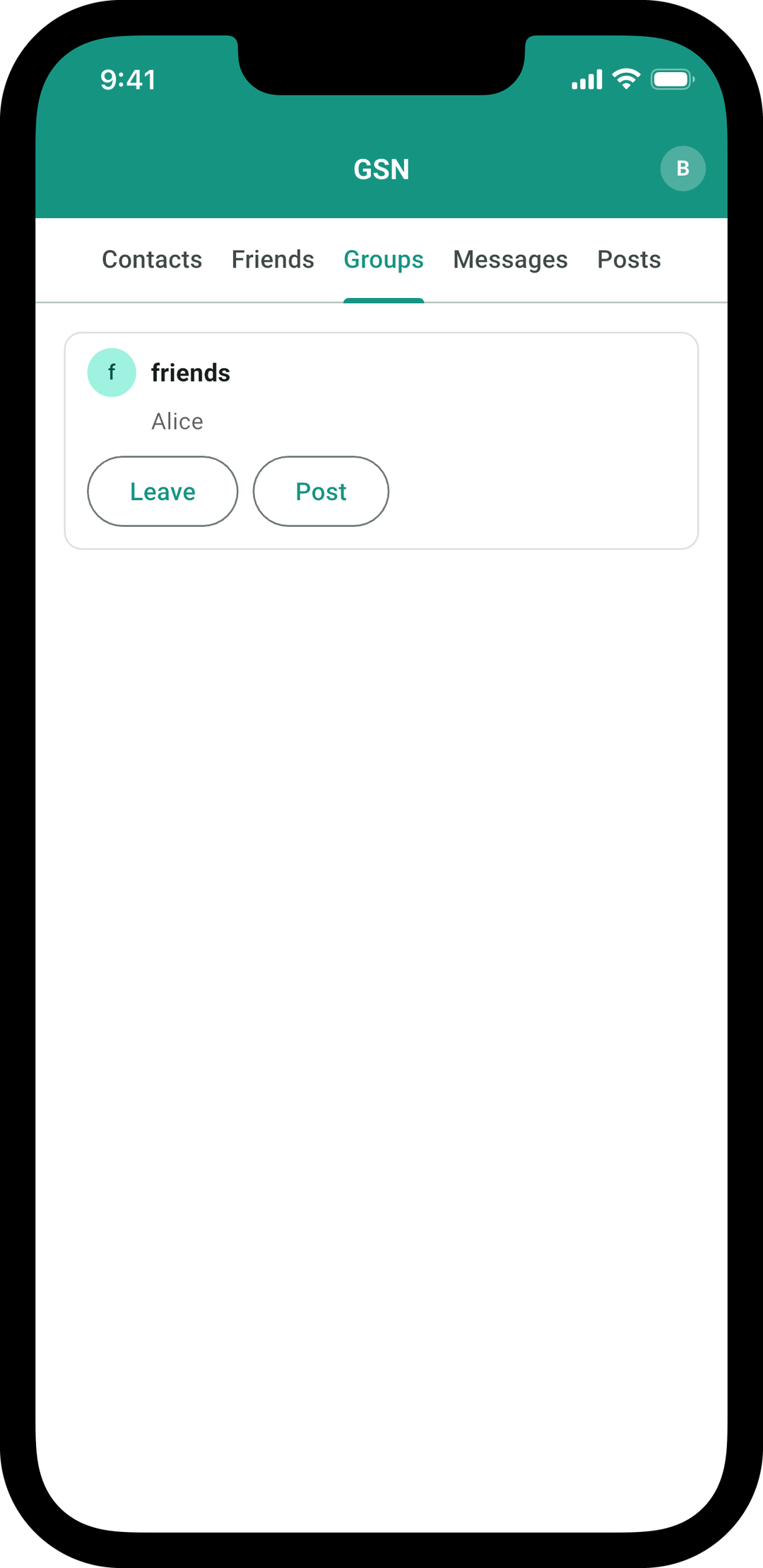}
\caption{The GSN app, on Bob's phone: Alice asks Bob to join her group \emph{friends}, and Bob answers yes or no (Groups:\ref{cl:joining}); the group \emph{friends}, led by Alice (Groups:\ref{cl:group}), which Bob may leave or post to (Groups:\ref{cl:joining} and Groups:\ref{cl:group-message}).}
\label{fig:gsn-app}
\end{figure}

The GSN app is a mini-app of the grassroots app, compiled down the cascade from Groups; Figure~\ref{fig:gsn-app} shows it realising Groups:\ref{cl:group}, Groups:\ref{cl:joining} and Groups:\ref{cl:group-message}.

\section{A Social Graph with Accountable AI Agents}
\label{sec:agents}

This section presents a grassroots contract~\cite{cardelli2020digital,shapiro2026formalising} for a social graph in which AI agents take part, each with a principal and a developer accountable for it, and shows which parts of the second risk of Section~\ref{sec:ai-risks} the contract excludes and which it makes answerable.  The contract has three parts: the grassroots social graph with Knowledge (Section~\ref{sec:gsg}) and Guardians, for guardians and their wards, both shared with the child-safe social network~\cite{shapiro2026cssn}, whose third part is for parents and children; and the AI-agents extension, the third part here; Section~\ref{sec:provenance} adds the social network of Section~\ref{sec:gsn} and its provenance.

The running example is Alice's AI agent Alf befriending Bob, Acme Ltd being Alf's developer.  It rests on three clauses: \enf~Alf and Bob, who know each other, may become friends by an act of Alf, Alice and Bob, provided Alice and Bob are friends (Guardians:\ref{cl:ward-person}); \enf~every act of Alf requires Alice's attestation, signed over Alf's public key, that Alf runs code signed by its developer, Acme Ltd (Agents:\ref{cl:ai-acts}); \und~Acme Ltd is accountable for Alf's defects, and Alice for operating Alf recklessly (Agents:\ref{cl:accountability}).

\subsection{A Social Contract for Guardians and Wards}
\label{sec:agents-contract}

Guardians is a contract for guardians and their wards~\cite{shapiro2026formalising}; Agents instantiates it for AI agents, and the child-safe social network for children~\cite{shapiro2026cssn}.

\begin{contractbox}{Guardians~\cite{shapiro2026formalising}}
Guardians extends and restricts Friendship.\par
A \emph{relation} is a friendship or the membership of a group.  Bob \emph{trusts} Carol, a ward of Alice, when he trusts Alice.  The \emph{decision rule}, a parameter of Guardians, says who decides for a ward and how.  A ward takes an act of a party by those who decide for it.\par
Alice and Bob are persons, Carol and Dave wards.
\begin{enumerate}[label=\arabic*.,ref=\arabic*]
\item \enf~\emph{Enter:}\label{cl:creation}  Alice may create a ward, Carol, and establish Carol's identity by choosing Carol's key pair.  Alice is then the creator, a guardian and a friend of Carol.
\item \enf~\emph{Guardian decisions.}\label{cl:joint}  \emph{Make guardian:} Those who decide for Carol may make Bob, a friend of a guardian of Carol, a guardian and a friend of Carol, with the consent of Bob.  \emph{Resign:} Bob may end his guardianship of Carol.  \emph{End guardian:} Those who decide for Carol, other than Bob, may end Bob's guardianship.  \emph{Guardian unfriend:} Bob may end his friendship with Carol, which ends his guardianship.  Carol may not end a friendship with her guardian.
\item \enf~\emph{Ward and person.}\label{cl:ward-person}  \emph{Ward acquaint:} Carol, a ward of Alice, and Bob may jointly come to know each other.  \emph{Ward befriend:} Carol may enter a relation with Bob with the consent of Bob and Alice, provided Carol and Bob know each other and Alice and Bob are friends.  \emph{Ward end, Unfriend, Ward unfriend:} Carol, Bob or Alice may end the relation of Bob and Carol.
\item \enf~\emph{Wards.}\label{cl:ward-ward}  \emph{Wards acquaint:} Carol, a ward of Alice, and Dave, a ward of Bob, may jointly come to know each other.  \emph{Wards befriend:} Carol may enter a relation with Dave with the consent of Dave, Alice and Bob, provided Carol and Dave know each other, Alice knows Dave and is a friend of Bob, Alice and Bob possibly one person.
\item \enf~\emph{Retire:}\label{cl:end-supervision}  Those who decide for Carol may retire Carol; every guardianship and relation of Carol then ends and Carol can take no act.
\end{enumerate}
\end{contractbox}

A ward may also be a business, whose guardians are its officers; we defer specifying how officers jointly exercise control over a business to future work.

An AI agent is made by a person, its \emph{developer}, and is operated by its guardians, among them the person who created it, its \emph{principal}.  Accountability for an AI agent is divided between the two as it is for a car between its manufacturer and its driver: the manufacturer is accountable for manufacturing defects, and the driver for reckless driving.  So the developer signs the code of the AI agent, and its principal attests, in a speech act signed over the AI agent's public key, that the AI agent runs that code, and every act of the AI agent requires the attestation.

\begin{contractbox}{Agents}
Agents instantiates Guardians with the guardians deciding jointly, an AI agent for a ward and a principal for a creator, composes it with Forwarding~\cite{shapiro2026formalising}, and restricts, strengthens and extends the composition.\par
The \emph{developer} of an AI agent is the signer of its code.\par
Alf is an AI agent.
\begin{enumerate}[label=\arabic*.,ref=\arabic*]
\item \und~\emph{AI developer and guardian accountability.}\label{cl:accountability}  The developer of Alf will be accountable for Alf's defects.  Every guardian of Alf will be accountable for operating it recklessly.
\item \enf~\emph{AI acts.}\label{cl:ai-acts}  \emph{Enter:} The principal of Alf attests at its creation (Guardians:\ref{cl:creation}), signed over Alf's public key, that Alf runs code signed by its developer.  Every act of Alf requires this attestation and names its developer and its principal.
\end{enumerate}
\iffull
\pragmas{Interface}
\emph{Known AI agents:} the attestation records Alice holds.\par
\emph{Enter:} \texttt{Alice [Enter] \textless{}name\textgreater{}}\par
\hspace*{1.5em}\emph{Carol:} \texttt{Alice asks to be your principal.}\par
\emph{Guardian decisions:} \texttt{Decider [End guardian] \textless{}person\textgreater{}}\par
\hspace*{1.5em}\emph{Codecider:} \texttt{Decider asks to end Bob's guardianship of your AI agent Carol.}\par
\emph{Retire:} \texttt{Decider [Retire]}\par
\hspace*{1.5em}\emph{Codecider:} \texttt{Decider asks to retire your AI agent Carol.}\par
\emph{AI acts:} \texttt{Decider [Make guardian] \textless{}person\textgreater{}}\par
\hspace*{1.5em}\emph{Bob:} \texttt{Decider asks you to be a guardian of the AI agent Carol.}\par
\hspace*{1.5em}\emph{Codecider:} \texttt{Decider asks that Bob be a guardian of your AI agent Carol.}\par
\hspace*{1.5em}\texttt{Bob [AI agent acquaint] \textless{}AI agent\textgreater{}}\par
\hspace*{1.5em}\emph{Carol:} \texttt{Bob wishes that you know each other.}\par
\hspace*{1.5em}\texttt{Alice [AI agent befriend] \textless{}AI agent\textgreater{}}\par
\hspace*{1.5em}\emph{Carol:} \texttt{Alice asks that you and Bob befriend.}\par
\hspace*{1.5em}\emph{Bob:} \texttt{Alice asks that you and her AI agent Carol befriend.}\par
\hspace*{1.5em}\texttt{Alice [AI agent unfriend] \textless{}AI agent\textgreater{} \textless{}person\textgreater{}}\par
\hspace*{1.5em}\texttt{Carol [AI agent end]}\par
\hspace*{1.5em}\texttt{Carol [AI agents acquaint] \textless{}AI agent\textgreater{}}\par
\hspace*{1.5em}\emph{Dave:} \texttt{Carol wishes that you know each other.}\par
\hspace*{1.5em}\texttt{Alice [AI agents befriend] \textless{}AI agent\textgreater{}}\par
\hspace*{1.5em}\emph{Carol:} \texttt{Alice asks that you and Dave befriend.}\par
\hspace*{1.5em}\emph{Dave:} \texttt{Alice asks that you and Carol befriend.}\par
\hspace*{1.5em}\emph{Bob:} \texttt{Alice asks that Carol and your AI agent Dave befriend.}\par
\hspace*{1.5em}\texttt{Alice [AI agents of one guardian befriend] \textless{}AI agent\textgreater{} \textless{}other AI agent\textgreater{}}\par
\hspace*{1.5em}\emph{Carol:} \texttt{Alice asks that you and Dave befriend.}\par
\hspace*{1.5em}\emph{Dave:} \texttt{Alice asks that you and Carol befriend.}\par
\hspace*{1.5em}\texttt{Alice [Sign] \textless{}text\textgreater{} [Forward]}\par
\hspace*{1.5em}\texttt{Alf [Agent sign] \textless{}text\textgreater{} [Agent forward]}\par
\fi
\end{contractbox}

\medskip

In Agents the wards are AI agents and their guardians the persons who operate them.  The principal's will is given once, when they create the AI agent; after that the AI agent acts on its own, and a guardian takes part in each friendship it enters (Guardians:\ref{cl:ward-person} and Guardians:\ref{cl:ward-ward}): an AI agent comes to know an agent by an act of the two, and befriends one only with a guardian.  The principal signs the attestation once too, when they create the AI agent, and the AI agent presents it to each agent it comes to know, who records it (Agents:\ref{cl:ai-acts}).  The guardians of an AI agent may make a friend of one of them a guardian of it (Guardians:\ref{cl:joint}), and may retire it (Guardians:\ref{cl:end-supervision}).  A guardian signs the undertaking to keep the contract for their ward~\cite{shapiro2026formalising}.

\begin{figure*}[t]
\centering
\includegraphics[width=0.22\textwidth,trim=0 460 0 0,clip]{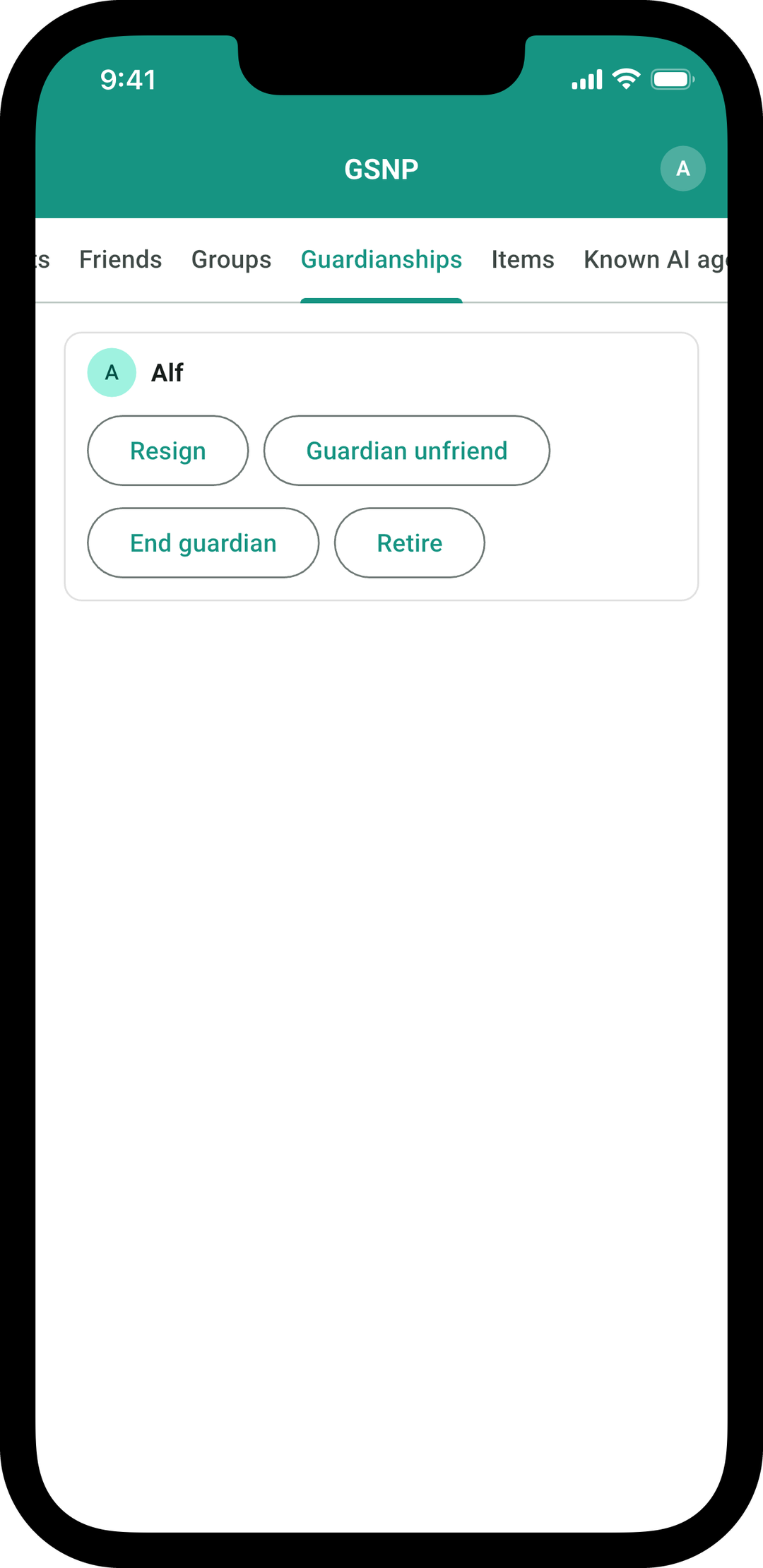}\hfill
\includegraphics[width=0.22\textwidth,trim=0 460 0 0,clip]{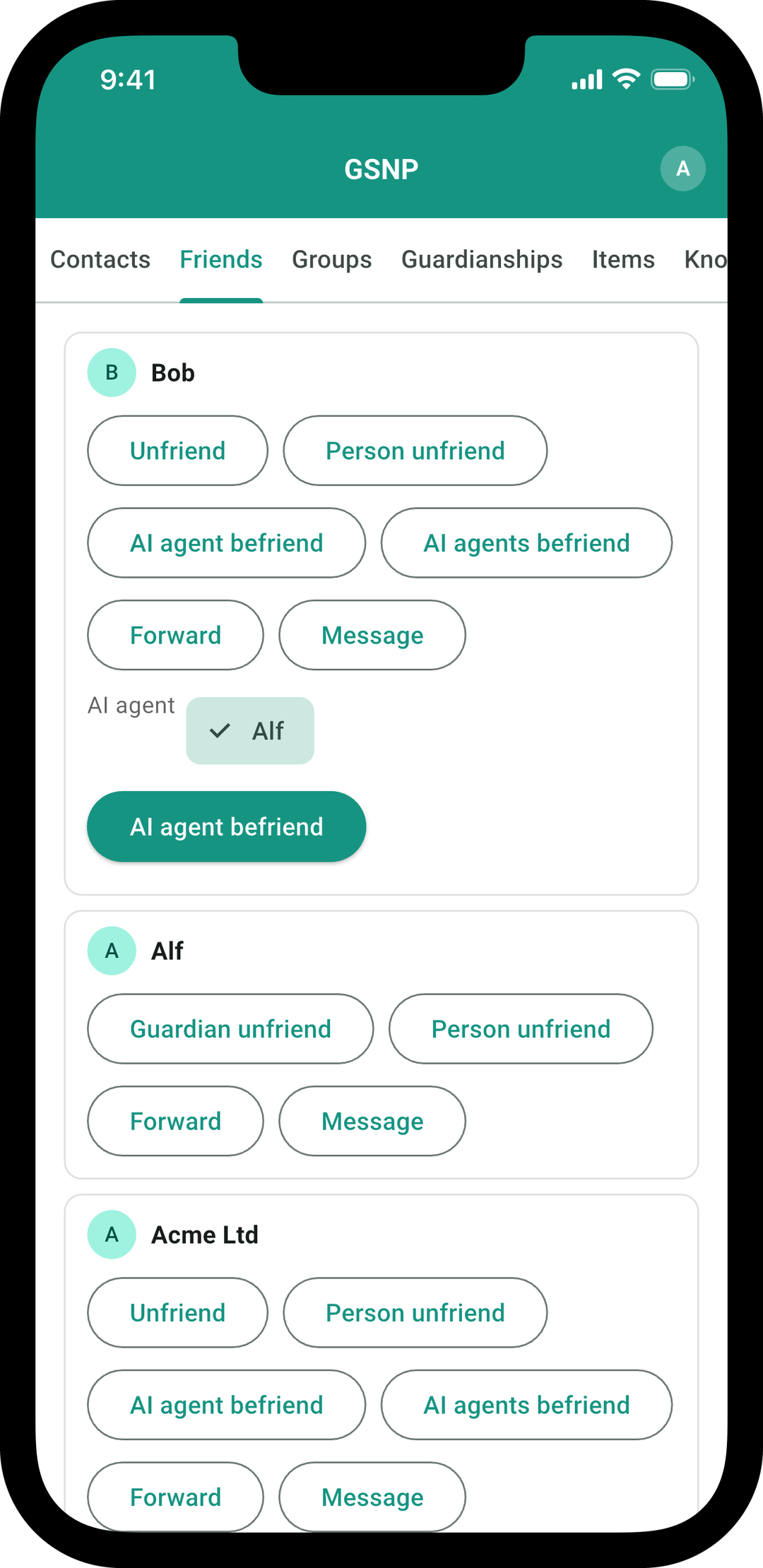}\hfill
\includegraphics[width=0.22\textwidth,trim=0 460 0 0,clip]{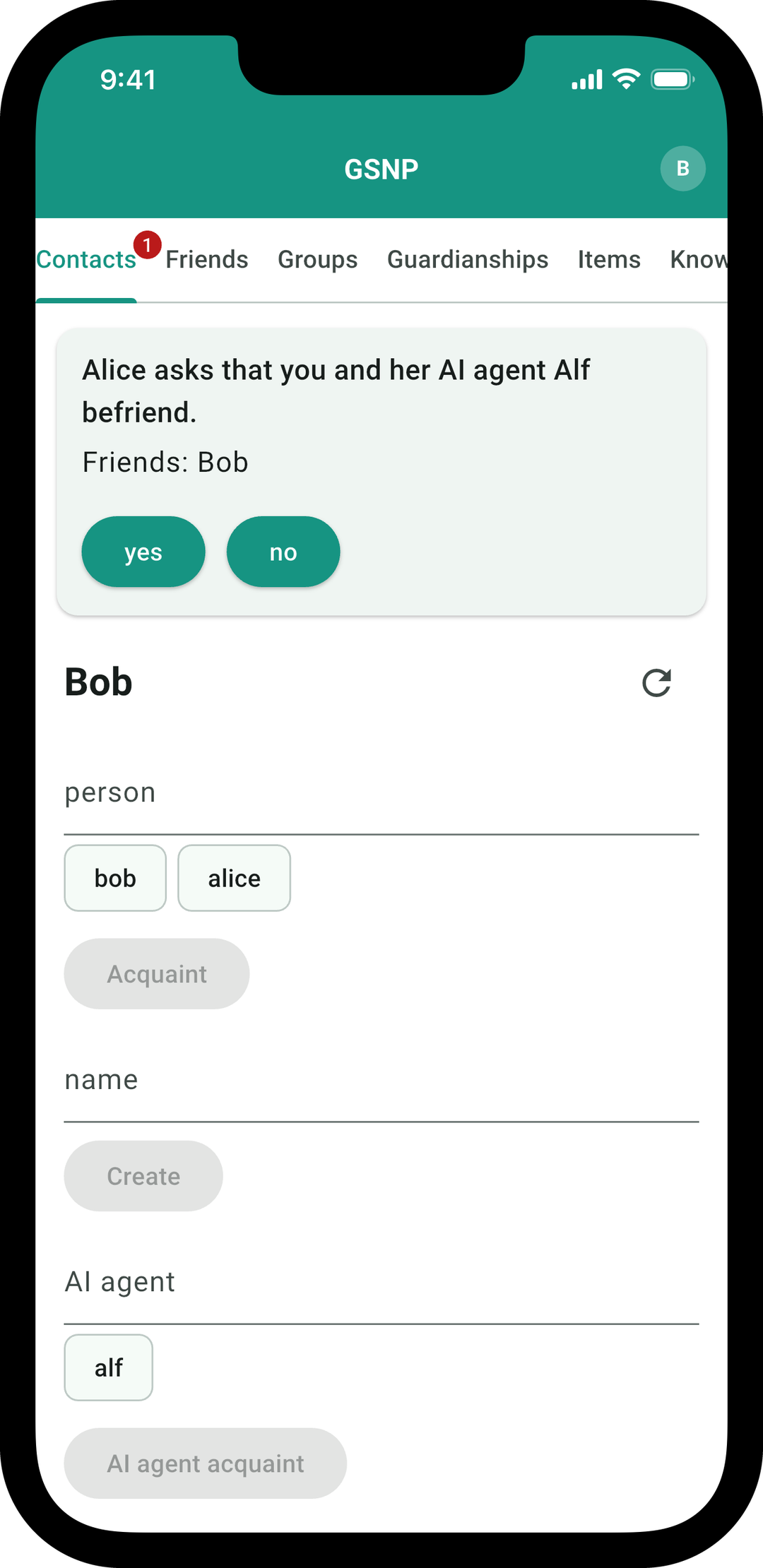}\hfill
\includegraphics[width=0.22\textwidth,trim=0 460 0 0,clip]{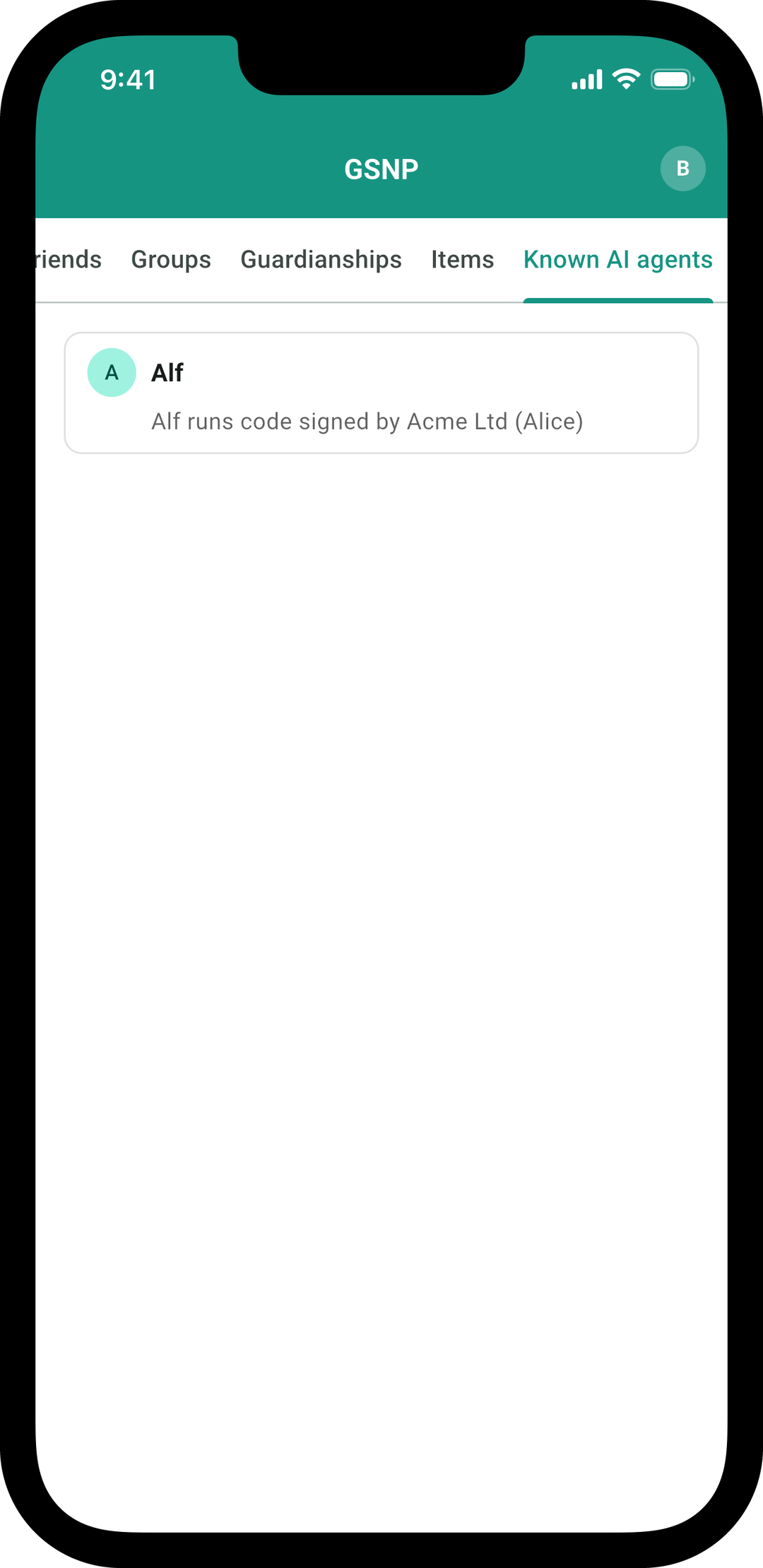}
\caption{Accountable AI agents, in the GSNP app: Alice's AI agent Alf, with the acts of its guardian, Retire among them (Guardians:\ref{cl:joint} and Guardians:\ref{cl:end-supervision}); Alice proposes that Alf befriend Bob (Guardians:\ref{cl:ward-person}); Bob is asked to befriend Alf, an AI agent of Alice (Guardians:\ref{cl:ward-person}); the AI agents Bob knows, Alf with Alice's attestation that it runs code signed by Acme Ltd (Agents:\ref{cl:ai-acts}).}
\label{fig:gai-app}
\end{figure*}

Figure~\ref{fig:gai-app} shows the contract on a person's phone, in the GSNP app of Section~\ref{sec:provenance}: a person creates an AI agent, attesting that it runs code signed by its developer, approves each friendship of the AI agent, and may retire it; every agent an AI agent knows holds its principal's attestation, which names its developer.

\subsection{The speech-act schemas}
\label{sec:agents-transactions}

The clauses of the contract compile, by their Logic pragmas, into speech-act schemas~\cite{shapiro2026formalising}, listed in Appendix~\ref{app:transactions}, which compile into volition-guarded multiagent atomic transactions~\cite{lewis2026volitional}, each a change to the states of its participants that occurs only if the agents guarding it are willing; the protocol realising the contract is the protocol over them~\cite{shapiro2026formalising,lewis2026volitional}, \full{recalled in Appendix~\ref{app:foundations}}{of~\cite{lewis2026volitional}}.  A schema is written with its head before the arrow and its body after.  The first argument of an atom is its holder, the agent in whose state it is; the body is what the act requires of the machines and, negated, forbids; the head is what it adds and, negated, deletes; and the guard names the agents who must will it.  Every role requires $\mathit{identity}$, a guarding role of a person $\mathit{person}$, and a guarding role of an AI agent that it have a guardian, which are not shown.  An agent $p$ is a \emph{person} when it holds $\mathit{person}(p)$, and an agent $r$ is an \emph{AI agent} when it holds $\mathit{ai\_agent}(r)$; its \emph{principal} is the $p$ holding $\mathit{principal}(p,r)$, its \emph{guardians} the $g$ with $\mathit{has\_guardian}(r,g)$, each holding $\mathit{guardian}(g,r)$, its \emph{attestation} the $y$ with $\mathit{attestation}(r,y)$, and an agent $s$ holds a \emph{record} of it as $\mathit{attestation\_record}(s,r,y)$; the predicates are listed in Table~\ref{tab:local-state} (Appendix~\ref{app:transactions}).

\begin{example}[An AI agent coming to know a person, and befriending them]\label{ex:agent-friendship}
The two schemas below realise Guardians:\ref{cl:ward-person} for an AI agent Carol, a guardian Alice of Carol and a person Bob: Carol, who has a guardian, and Bob come to know each other by an act of the two, Bob recording Carol's attestation $\mathit{Proof}$ (Agents:\ref{cl:ai-acts}), and then, Alice and Bob being friends, may become friends by an act of the three; Carol, Bob, and a guardian of Carol may end the friendship by the other three schemas of Guardians:\ref{cl:ward-person} (Appendix~\ref{app:transactions}).
\begin{contractbox}{Guardians:3: Speech-Act Schemas}
\raggedright
\begin{enumerate}[label=\arabic*.]
\item[3.]
{$\mathit{knows}(\Carol, \Bob)$, $\mathit{knows}(\Bob, \Carol)$, $\mathit{attestation\_record}(\Bob, \Carol, \mathit{Proof})$ $\leftarrow$}
\sasbody{$\mathit{ai\_agent}(\Carol)$, $\mathit{has\_guardian}(\Carol, \Alice)$, $\mathit{person}(\Bob)$, $\mathit{attestation}(\Carol, \mathit{Proof})$ guarded~by~$\Carol$, $\Bob$.}
\smallskip
\sashead{$\mathit{friend}(\Carol, \Bob)$, $\mathit{friend}(\Bob, \Carol)$ $\leftarrow$}
\sasbody{$\mathit{ai\_agent}(\Carol)$, $\neg\mathit{retired}(\Carol)$, $\mathit{has\_guardian}(\Carol, \Alice)$, $\mathit{person}(\Bob)$, $\mathit{knows}(\Carol, \Bob)$, $\mathit{knows}(\Bob, \Carol)$, $\mathit{friend}(\Alice, \Bob)$, $\mathit{friend}(\Bob, \Alice)$, $\neg\mathit{friend}(\Carol, \Bob)$, $\neg\mathit{friend}(\Bob, \Carol)$, $\mathit{attestation}(\Carol, \mathit{Proof})$ guarded~by~$\Carol$, $\Bob$, $\Alice$.}
\end{enumerate}
\end{contractbox}
\end{example}

For example, let Alice and Bob be persons and Alf Alice's AI agent.  Alice and Bob establish their identities; Alice signs $y$, her attestation that Alf, by its public key, runs the code its developer signed, and creates Alf with it; Alice and Bob come to know each other; Alf and Bob come to know each other, and Bob records Alf's attestation; Alice and Bob befriend; Alf befriends Bob, with Alice willing; Alf signs an item $x$ and sends it to Bob; and Alice retires Alf, which ends Alf's friendship with Bob.  \full{The atoms each act adds and deletes are in Table~\ref{tab:example} (Appendix~\ref{app:transactions}), Bob's}{Bob's} record and the item \full{remaining}{remain} with him after the friendship ends.

\subsection{Properties}
\label{sec:agents-properties}

This section states what the schemas enforce about an AI agent and its principal, with the argument for it, and that the protocol is volitionally grassroots, proved in Appendix~\ref{app:proofs}.

\mypara{The standing of a principal} Consider any configuration reached by a run of the protocol (Appendix~\ref{app:transactions}) and an AI agent $r$ in it, with $p$ holding $\mathit{principal}(p,r)$.
\begin{enumerate}[label=(\arabic*),ref=\arabic*]
\item \emph{Creation.}\label{st:guardianship}  The identity of $r$ was established by an act of Guardians:\ref{cl:creation} with $p$ and $r$ in its roles, guarded by both, for an item $y$ that $p$ holds and authored, and $\mathit{ai\_agent}(r)$, $\mathit{attestation}(r,y)$, $\mathit{principal}(p,r)$ and $\mathit{attestation\_record}(p,r,y)$ hold in every configuration since; $r$ holds $\mathit{has\_guardian}(r,g)$ if and only if $g$ holds $\mathit{guardian}(g,r)$; and every guardian $g$ of $r$ is a person and holds $\mathit{attestation\_record}(g,r,y)$.
\item \emph{Whom $r$ knows and befriends.}\label{st:friends}  Every $s$ other than $p$ and $r$ with $\mathit{knows}(r,s)$ became a guardian of $r$ by an act of Guardians:\ref{cl:joint} or came to know $r$ by an act of Guardians:\ref{cl:ward-person} or Guardians:\ref{cl:ward-ward}, and holds $\mathit{attestation\_record}(s,r,y)$; every $s$ with $\mathit{friend}(r,s)$ became a friend of $r$ on becoming a guardian of $r$, by an act of Guardians:\ref{cl:creation} or Guardians:\ref{cl:joint}, or is known to $r$ and became a friend of $r$ by an act of Guardians:\ref{cl:ward-person} or Guardians:\ref{cl:ward-ward} guarded by a guardian of $r$.
\item \emph{The power to end each.}\label{st:power}  For every guardian $g$ of $r$ and every friend $s$ of $r$ that is not a guardian of $r$, the act of Guardians:\ref{cl:ward-person} by which $g$ ends the friendship of $r$ and $s$ is machine-enabled and guarded by $g$; and while $r$ has a guardian, the act of Guardians:\ref{cl:end-supervision} by which the guardians of $r$ retire it, ending every friendship and guardianship of $r$, is machine-enabled and guarded by the guardians of $r$.
\item \emph{An AI agent with no guardian.}\label{st:retirement}  If $r$ has no guardian then $r$ has left, and $\mathit{ai\_agent}(r)$, $\mathit{attestation}(r,y)$ and $\mathit{principal}(p,r)$ hold; a retired AI agent has no guardian.
\end{enumerate}
Each holds by inspection of the schemas.  (1) Only the schema of Guardians:\ref{cl:creation} adds $\mathit{ai\_agent}$, $\mathit{principal}$ and $\mathit{attestation}$, and no schema deletes them or an atom of $\mathit{attestation\_record}$, so $\mathit{ai\_agent}(r)$, $\mathit{attestation}(r,y)$, $\mathit{principal}(p,r)$ and $\mathit{attestation\_record}(p,r,y)$ were added by one act of it, which requires $\mathit{item}(p,y)$ with $\mathit{author}(y) = p$, forbids $\mathit{identity}(r)$ and is guarded by $p$ and $r$; before it $r$ held nothing, as every other schema requires $\mathit{identity}$ at each of its roles.  The schemas of Guardians:\ref{cl:creation} and Guardians:\ref{cl:joint} for a new guardian add $\mathit{has\_guardian}(r,g)$ at $r$ together with $\mathit{guardian}(g,r)$ and $\mathit{attestation\_record}(g,r,y)$ at $g$, where they require $\mathit{person}$, the second requiring $\mathit{attestation}(r,y)$ at $r$; those of Guardians:\ref{cl:joint} by which a guardianship ends and that of Guardians:\ref{cl:end-supervision} delete $\mathit{has\_guardian}(r,g)$ and $\mathit{guardian}(g,r)$ together; and no schema deletes $\mathit{person}$.  (2) The schemas adding $\mathit{knows}(r,\cdot)$ are those of Knowledge:\ref{cl:meeting} and Knowledge:\ref{cl:introduction}, which require $\mathit{person}$ at their roles, and $r$, holding $\mathit{ai\_agent}$, is not a person, a schema adding $\mathit{identity}$ adding $\mathit{person}$ or $\mathit{ai\_agent}$ and not both; that of Guardians:\ref{cl:creation}, which adds $\mathit{knows}(r,p)$; that of Guardians:\ref{cl:joint}, which adds $\mathit{knows}(r,g)$ for the new guardian $g$, with $\mathit{attestation\_record}(g,r,y)$ at $g$; and those of Guardians:\ref{cl:ward-person} and Guardians:\ref{cl:ward-ward} for knowing, which require $\mathit{attestation}(r,y)$ and add $\mathit{attestation\_record}(\cdot,r,y)$ at the agent coming to know $r$.  The schemas adding $\mathit{friend}(r,\cdot)$ are that of Friendship:\ref{cl:friendship}, which requires $\mathit{person}$ at its roles; those of Guardians:\ref{cl:creation} and Guardians:\ref{cl:joint}, which add a guardian; and those of Guardians:\ref{cl:ward-person} and Guardians:\ref{cl:ward-ward} for a relation, which require that $r$ know the new friend and are guarded by a guardian of $r$.  (3) The bodies of the two acts hold: $r$ holds $\mathit{has\_guardian}(r,g)$; $g$ is a person, and so is every guardian of $r$, each holding $\mathit{guardian}(\cdot,r)$, by (1); $s$ does not hold $\mathit{guardian}(s,r)$; every friend of $r$ holds $\mathit{friend}$ of $r$, the schemas adding and deleting the two atoms of a friendship together; and $r$, having a guardian, is not retired, the schema of Guardians:\ref{cl:end-supervision} being the one that adds $\mathit{retired}$ and clearing $\mathit{has\_guardian}$ at $r$, and that of Guardians:\ref{cl:joint} for a new guardian forbidding $\mathit{retired}$.  (4) Every role $r$ guards in a schema other than that of Guardians:\ref{cl:creation}, which forbids $\mathit{identity}$ there, requires $\mathit{person}$, which $r$ does not hold, or an atom of $\mathit{has\_guardian}$ at $r$: the acts of an AI agent in Guardians:\ref{cl:ward-person} and Guardians:\ref{cl:ward-ward}, and its signing, forwarding and messaging (Agents:\ref{cl:ai-acts} and Provenance Groups:\ref{cl:agent-message}), ask that it have a guardian, and every other guarded act is a person's; and the one schema adding $\mathit{has\_guardian}$ at $r$ once it has an identity, that of Guardians:\ref{cl:joint}, requires one there.  So once $r$ holds no atom of $\mathit{has\_guardian}$ no act $r$ guards is machine-enabled: $r$ has left; $\mathit{ai\_agent}(r)$, $\mathit{attestation}(r,y)$ and $\mathit{principal}(p,r)$ hold by (1); and the schema of Guardians:\ref{cl:end-supervision} clears $\mathit{has\_guardian}$ at $r$.

Part~(1) is why what the contract says of an AI agent is enforced rather than attested: an agent the contract records as an AI agent of $p$ cannot cease to be one by any act, its own or another's, and by part~(4) it takes no act without a guardian.  Retirement ends every guardianship and friendship of the AI agent (Guardians:\ref{cl:end-supervision}): the retired AI agent takes no act; the record of its principal stays, for its past acts.

\mypara{Grassroots} The protocol realising the contract is volitionally grassroots (Proposition~\ref{prop:grassroots}, Appendix~\ref{app:proofs}, by the theorem for contracts with wards of~\cite{shapiro2026formalising}): two persons can always befriend, an AI agent with a guardian can always come to know a person or another AI agent with a guardian, and befriend them with the guardians willing, and through befriending any agent can come to interact with any other; two groups that have formed apart, each with an agent that has not left, can always coalesce, and only by an act an agent of each wills.  An AI agent with no guardian has left.

\subsection{What the contract excludes and what it makes answerable}
\label{sec:agents-modalities}

A clause is enforced, attested or undertaken (Section~\ref{sec:gsc})~\cite{shapiro2026formalising}: no act of agents can breach an enforced clause; a breach of an attested clause leaves non-repudiable evidence against the agent in breach in another agent's hands; and an undertaken clause is kept outside the contract, the agent's signed undertaking to keep the contract, held by every agent they interact with, being its record, so a breach of an attested or an undertaken clause can be taken to court.  The second risk of Section~\ref{sec:ai-risks} has five parts, and we take them in turn.

\mypara{Anonymity} A person signs their acts with the private key of their genuine digital identity (Knowledge:\ref{cl:identity}); an AI agent's key pair is chosen by its principal (Guardians:\ref{cl:creation}), and every act of an AI agent is signed with its private key and names its principal (Agents:\ref{cl:ai-acts}).  Both are enforced: an agent without an identity has no act of the contract available to it, and an act of the contract is a digital speech act, an utterance signed with the key of the agent taking it~\cite{golike2026digital,shapiro2026formalising}.

\mypara{Masquerading as a person} An AI agent is recorded as one on its creation (Guardians:\ref{cl:creation} and Agents:\ref{cl:ai-acts}), is a ward from its first act and remains one (Section~\ref{sec:agents-properties}(\ref{st:guardianship})), every agent it comes to know records it with its principal's attestation, and every friendship of it is an act of a guardian of it (Guardians:\ref{cl:ward-person}, Guardians:\ref{cl:ward-ward} and Agents:\ref{cl:ai-acts}), so every agent an AI agent knows, and so every agent it interacts with, holds a record of it with its principal's attestation, which names it as an AI agent and names its principal (Section~\ref{sec:agents-properties}(\ref{st:friends})): enforced.  What the contract cannot determine is whether an agent that chose its own key pair is a person: no act of the contract can tell a person's act from a program's taken under that person's identity.  Such a program is not an AI agent of the contract; its acts are signed with the person's private key and are that person's acts, for which the person is accountable (Knowledge:\ref{cl:identity}).  A person's accountability for the acts signed with their private key is undertaken: the person's signed undertaking, held by every agent they interact with, is its record.

\mypara{Unknown masters} The principal of an AI agent is a person with a genuine digital identity, the author of the attestation that the AI agent holds and each of its acts requires, and a guardian of it, a person, takes part in each of its friendships (Guardians:\ref{cl:ward-person}, Guardians:\ref{cl:ward-ward} and Agents:\ref{cl:ai-acts}): enforced.

\mypara{Unknown goals and restraints} The contract does not determine what an AI agent is for or what its principal lets it do.  It determines who is accountable for it: its developer for its defects, and every guardian of it for operating it recklessly (Agents:\ref{cl:accountability}).  The clause is undertaken, each guardian's signed undertaking being held by every agent they interact with.  Agents:\ref{cl:ai-acts} is enforced: the principal signs the attestation when they create the AI agent, and every act of the AI agent requires it, every agent the AI agent knows holding a record of it (Section~\ref{sec:agents-properties}(\ref{st:guardianship}) and~(\ref{st:friends})).  The attestation does not establish what code the AI agent runs; an attestation of code the AI agent does not run is evidence against the principal that they cannot repudiate.

\mypara{Accountability and culpability} Every act of an AI agent is signed with its key and requires its principal's attestation, which names its principal (Agents:\ref{cl:ai-acts}), and the agent it is addressed to holds it: non-repudiable evidence attributing the act to the AI agent and the AI agent to its principal~\cite{golike2026digital}.  Harm done by an AI agent under this contract is harm done by an identified agent, whose guardians are accountable for operating it recklessly and whose developer is accountable for its defects (Agents:\ref{cl:accountability}).

Hence on a platform realising this contract there are no anonymous agents, natural or computational, and so there are no anonymous AI agents: an AI agent acts under an identity of its own, recorded as an AI agent and naming its principal, or a program acts under a person's identity and that person is accountable for its acts.

\section{A Social Network with Digital Provenance}
\label{sec:provenance}

This section presents Provenance Groups, the grassroots social network with provenance~\cite{shapiro2023gsn}, which instantiates Groups (Section~\ref{sec:gsn}) with items authored by their senders, composes it with the contract of Section~\ref{sec:agents}, and restricts and extends the composition, so that a message is an item of Forwarding~\cite{shapiro2026formalising}: an agent may create an item by signing it, and forward an item they hold to a friend, adding their signature to the signatures the item carries.  It then shows which parts of the third risk of Section~\ref{sec:ai-risks} the contract excludes and which it makes answerable.  An \emph{item} is a speech act: content signed by its author~\cite{golike2026digital}, and, when forwarded, signed over by each agent that forwarded it, in order.  The speech-act schemas of Provenance Groups are in the repository, and the protocol realising it is volitionally grassroots (Proposition~\ref{prop:gsnp-grassroots}, Appendix~\ref{app:proofs}).

\subsection{Extending the contract with provenance}
\label{sec:provenance-contract}

\begin{contractbox}{Forwarding~\cite{shapiro2026formalising}}
Forwarding extends Friendship.\par
The \emph{author} of an item is its first signer.\par
Alice and Bob are persons.
\begin{enumerate}[label=\arabic*.,ref=\arabic*]
\item \enf~\emph{Sign:}\label{cl:item-creation}  Alice may create an item by signing it.
\item \enf~\emph{Forward:}\label{cl:forwarding}  Alice may forward an item she holds to her friend Bob, adding her signature to the signatures the item carries.
\end{enumerate}
\end{contractbox}

\begin{contractbox}{Provenance Groups}
Provenance Groups instantiates Groups~\cite{shapiro2026formalising} with items authored by their senders, composes it with Agents, and restricts and extends the composition.
\begin{enumerate}[label=\arabic*.,ref=\arabic*]
\item \enf~\emph{Agent message:}\label{cl:agent-message}  An AI agent may send a message to a friend (Groups:\ref{cl:message}).
\item \enf~\emph{Post forward:}\label{cl:post-forward}  A member of a group, not its leader, may forward an item they hold to the group, adding their signature.
\end{enumerate}
\pragmas{Interface}
\emph{Agent message:} \texttt{Alice [Message]}\par
\hspace*{1.5em}\texttt{Bob [Post]}\par
\hspace*{1.5em}\texttt{Alice [Post own]}\par
\hspace*{1.5em}\texttt{Alf [Agent message]}\par
\emph{Post forward:} \texttt{Bob [Post forward]}\par
\end{contractbox}

\medskip
Provenance Groups adds provenance to the messages of Groups: a message to a friend (Groups:\ref{cl:message}) or to a group (Groups:\ref{cl:group-message}) is an item its sender authored.  An agent signs the content they create, as they sign every act (Knowledge:\ref{cl:identity} and Forwarding:\ref{cl:item-creation}): signing adds the item to its author's holdings; a message carries an item its sender authored, as they signed it; forwarding adds to the friend's holdings an item the forwarder holds, signed over by the forwarder (Forwarding:\ref{cl:forwarding}); and a leader delivers a member's item to the group by forwarding it, signed over by the leader; a member of a group, not its leader, may forward to it an item they hold, adding their signature (Provenance Groups:\ref{cl:post-forward}); an AI agent may send a message to a friend (Provenance Groups:\ref{cl:agent-message}), and takes no act of a group; so an item in an agent's hands is signed by its author and by every agent that forwarded it, in order, and carries its provenance inside it.  Alice may remove the signatures of an item with other software and send the result as her own; it is then an item Alice created, of which she is the author (Forwarding:\ref{cl:item-creation}), and she is accountable for it.  An agent holds an item by signing it or by receiving it from a friend, the leader of a group being a friend of each of its members (Groups:\ref{cl:joining}), and in no other way: content that reaches a person otherwise is not an item of the platform.  Accountability for the items of an AI agent is as in Agents:\ref{cl:accountability}.

\begin{figure*}[t]
\centering
\includegraphics[width=0.22\textwidth,trim=0 740 0 0,clip]{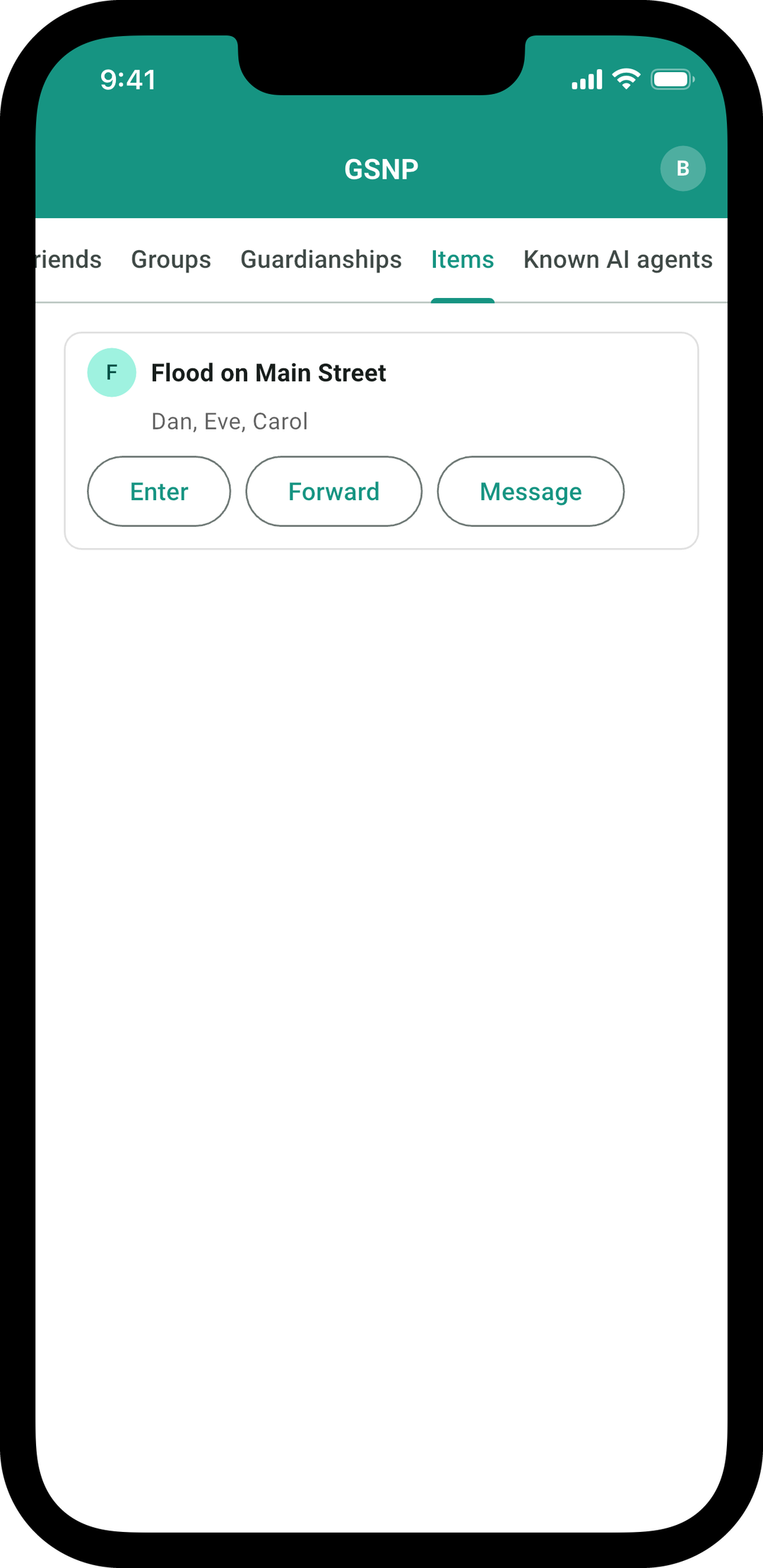}\hfill
\includegraphics[width=0.22\textwidth,trim=0 740 0 0,clip]{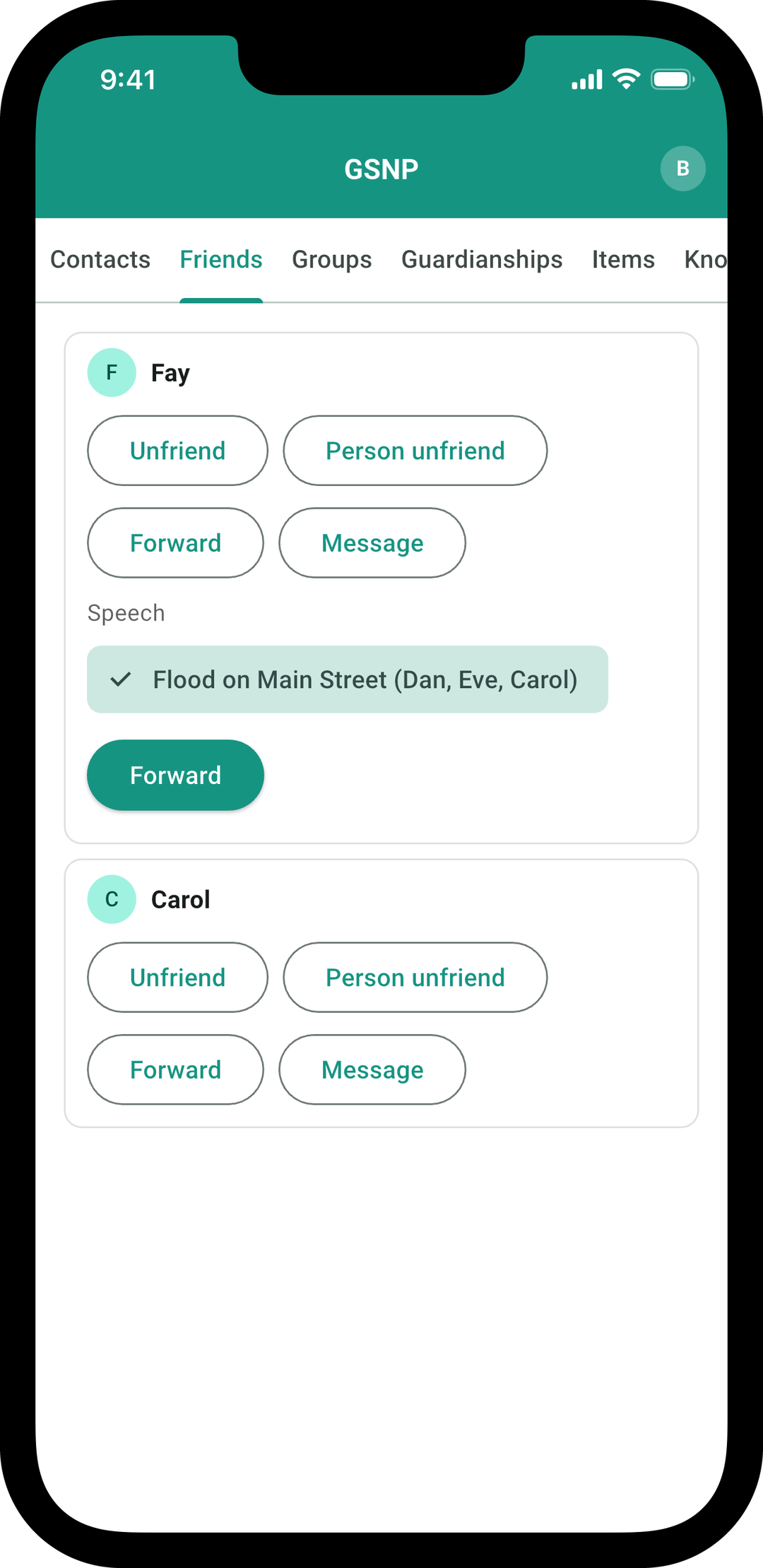}\hfill
\includegraphics[width=0.22\textwidth,trim=0 740 0 0,clip]{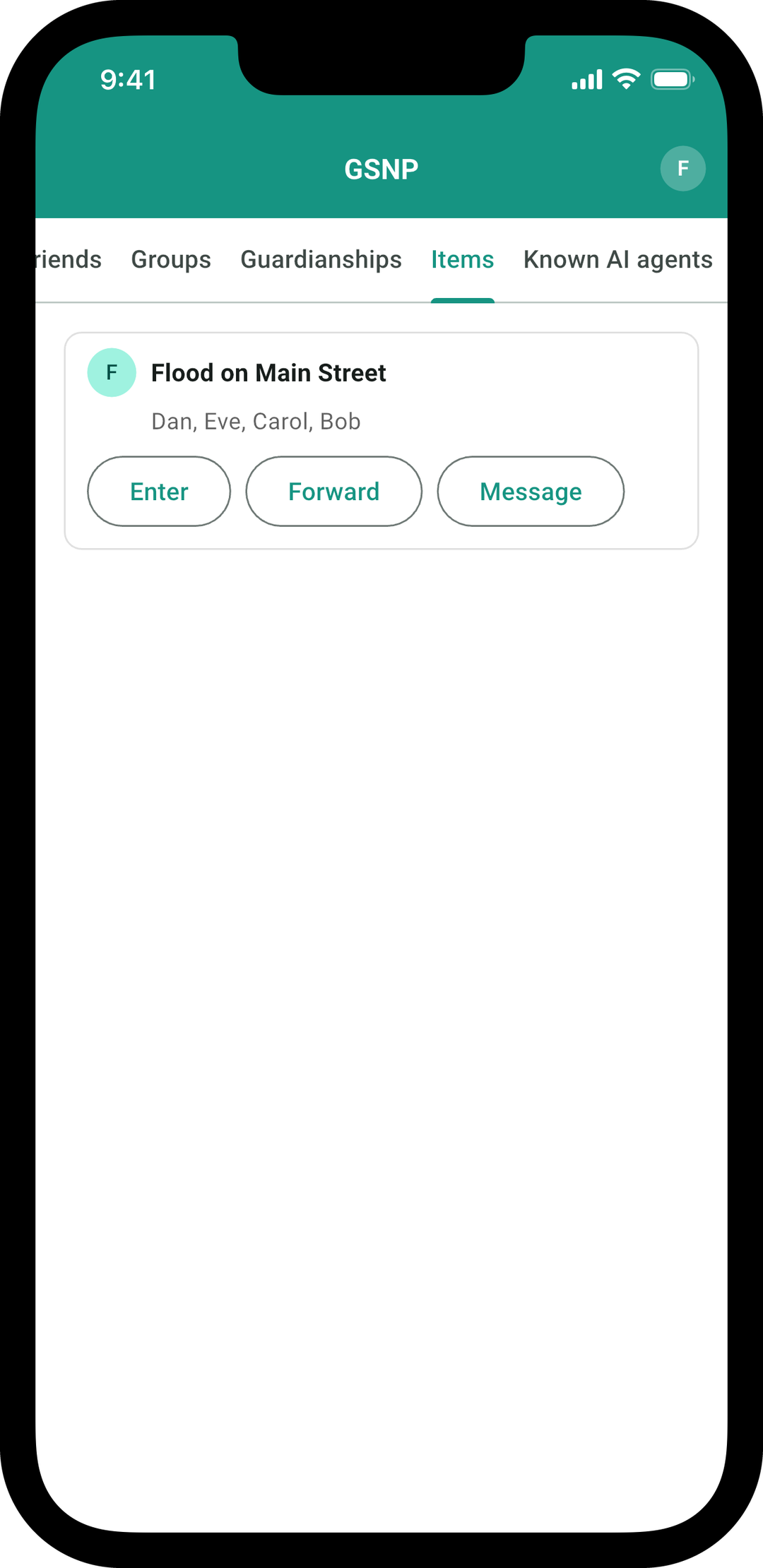}\hfill
\includegraphics[width=0.22\textwidth,trim=0 740 0 0,clip]{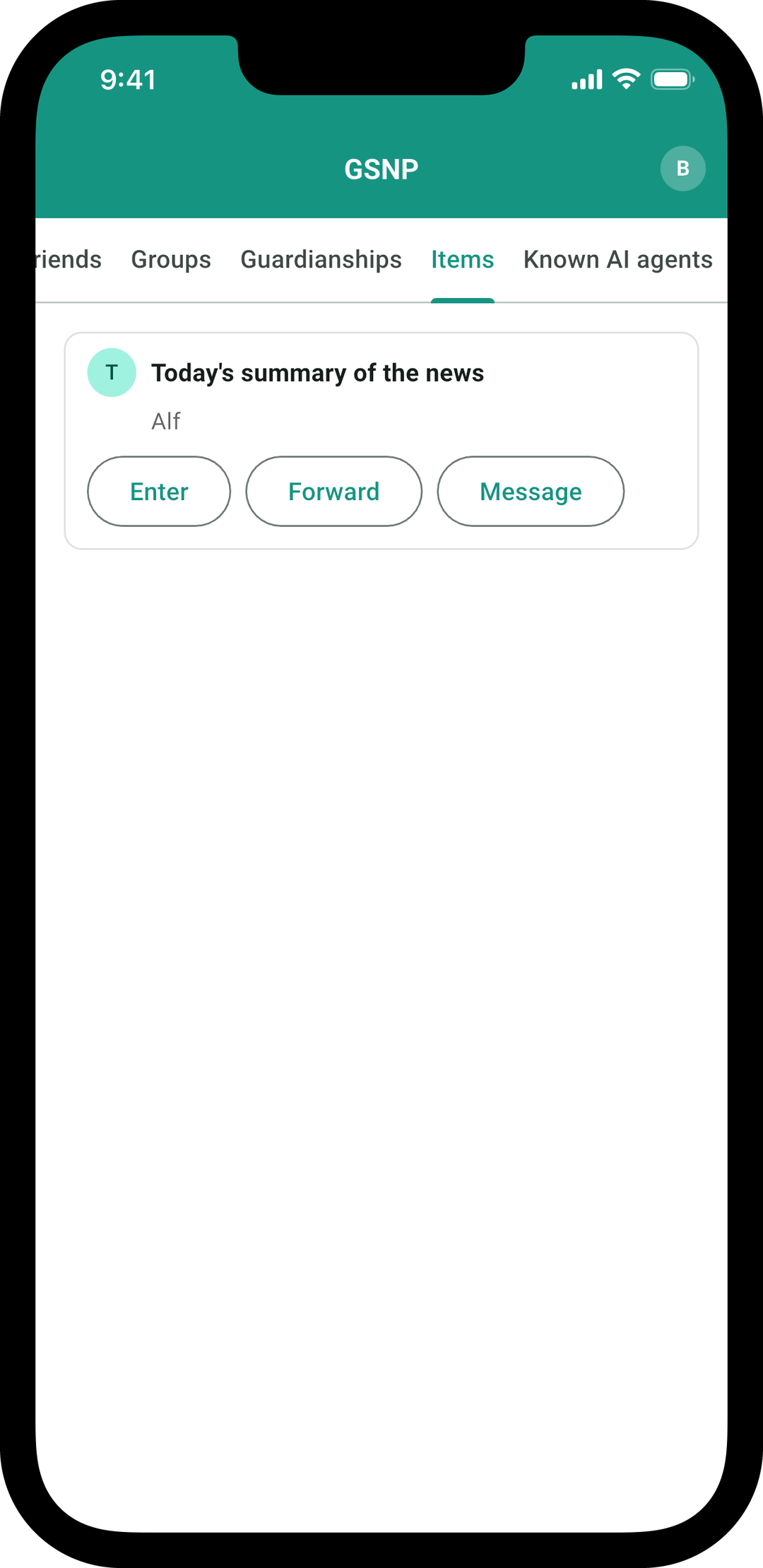}
\caption{The GSNP app: on Bob's phone, an item with its author and the agents that forwarded it (Forwarding:\ref{cl:forwarding}); Bob forwarding it to Fay (Forwarding:\ref{cl:forwarding}); on Fay's phone, the item with Bob among its forwarders; on Bob's phone, an item authored by the AI agent Alf (Agents:\ref{cl:ai-acts}).}
\label{fig:gsnp-app}
\end{figure*}

The GSNP app is the GSN app with provenance and with the AI agents of Section~\ref{sec:agents} (Figure~\ref{fig:gsnp-app}): every item shows its author and every agent that forwarded it (Forwarding:\ref{cl:forwarding}).

\subsection{What the contract excludes and what it makes answerable}
\label{sec:provenance-modalities}

The third risk of Section~\ref{sec:ai-risks} has three parts, and we take them in turn, with the three modalities of Section~\ref{sec:agents-modalities}.

\mypara{Anonymous content} Every item is signed by its author, an agent with a genuine digital identity who signs every act with their private key (Knowledge:\ref{cl:identity}), and an agent holds an item only by signing it or by receiving it from a friend, who authored it or signed it over (Forwarding:\ref{cl:forwarding}): enforced.  Every item an agent holds carries the signatures of its author and of each agent it passed through, and content signed by nobody is not an item of the platform.  Whether the content was produced by a person or by an AI is not what the signatures say; who is accountable for it is.

\mypara{Content people cannot tell from genuine content} The platform does not decide whether content is synthetic; it decides who signed it.  An item authored by an AI agent is signed by the AI agent, which every agent it knows holds recorded as an AI agent of its principal (Guardians:\ref{cl:ward-person} and Agents:\ref{cl:ai-acts}): a reader who knows the AI agent sees that the item is an AI agent's and whose; enforced.  A reader who does not know the AI agent sees its signature and the forwarders', each of whom received it from a friend (Forwarding:\ref{cl:forwarding}).  An item a person signs is that person's, whatever produced it, and the person is accountable for it (Knowledge:\ref{cl:identity}): a person who signs AI-generated content as their own is its author.  An item that purports to be a person's and is not carries that person's signature or it does not, and a signature cannot be forged~\cite{golike2026digital}; an item that depicts or quotes a person is its author's statement about them, and the author is accountable for it.  So a reader of a grassroots social network tells content with provenance from the rest by its signatures, which every item carries, and not by its look.

\mypara{Producers evading accountability} The signature of an item is non-repudiable evidence that its author produced it, held by every agent that holds the item, and each forwarder's signature over it is evidence that they passed it on: enforced (Forwarding:\ref{cl:forwarding}).  An agent who removes the signatures of an item they received and sends the result as their own is its author (Forwarding:\ref{cl:item-creation}), and is accountable for it.  The author of an item, a person or an AI agent with an accountable developer and principal (Agents:\ref{cl:accountability}), and each agent that forwarded it, are identified agents, and harm done by an item can be taken to court against them~\cite{golike2026digital,shapiro2026formalising}.

Hence on a platform realising this contract there is no anonymous content: every item carries the signatures of its author, a person accountable for it or an AI agent whose principal and developer are accountable for it, and of every agent that forwarded it.  Deepfake content may be produced outside the platform and signed into it, and it then has an author who is accountable for it; what the platform does not carry is content with no one behind it.

\section{Related Work}
\label{sec:related}

\mypara{AI risks} Taxonomies and surveys of AI risks include those of Weidinger et al.~\cite{weidinger2022taxonomy}, Bommasani et al.~\cite{bommasani2021opportunities} and Hendrycks et al.~\cite{hendrycks2023catastrophic}, the AI Risk Repository~\cite{slattery2026risk} and the International AI Safety Report~\cite{bengio2026safety}.  Algorithmic governance and regulation~\cite{danaher2017algorithmic,yeung2018algorithmic} and gradual disempowerment~\cite{kulveit2025disempowerment} bear on the first risk; personhood credentials~\cite{adler2024personhood}, social bots~\cite{ferrara2016rise}, visibility into AI agents~\cite{chan2024visibility}, the governance of agentic systems~\cite{shavit2023practices}, moral crumple zones~\cite{elish2019crumple} and meaningful human control~\cite{santonidesio2018meaningful} on the second; deepfakes~\cite{chesney2019deep,mirsky2021creation,nightingale2022faces,vaccari2020deepfakes} and reducing the risks of synthetic content~\cite{chandra2024synthetic} on the third; and AI assistants~\cite{gabriel2024ethics}, chatbot use~\cite{fang2025psychosocial} and malicious use~\cite{brundage2018malicious,marchal2024misuse} on the two risks not addressed here.  We address the three through the organisation of the platform.

\mypara{Grassroots platforms} Grassroots platforms~\cite{shapiro2023grassrootsBA,shapiro2024grassroots} are characterised against global platforms, centralised, decentralised and federated~\cite{shapiro2025characterising}, the decentralised ones exemplified by Bitcoin~\cite{nakamoto2008bitcoin} and Ethereum~\cite{ethereum}.  They are specified by atomic transactions~\cite{shapiro2025atomic} and volition-guarded multiagent atomic transactions~\cite{lewis2026volitional}, and defined by grassroots contracts~\cite{shapiro2026formalising}, digital social contracts~\cite{cardelli2020digital} meeting Proudhon's conditions~\cite{proudhon2004general}.  Their instances include grassroots social networks~\cite{shapiro2023gsn}, currencies~\cite{shapiro2024gc} and bonds~\cite{shapiro2026bonds}, and a child-safe social network~\cite{shapiro2026cssn}.  Their identities are genuine personal identifiers~\cite{shahaf2020genuine}, their communities grow sybil-resiliently~\cite{poupko2021building}, their content is digital speech acts~\cite{golike2026digital}, and their participants recover from the loss or compromise of a key or a device with the help of their friends~\cite{eitan2026securing}.

\mypara{The cascade} A contract is compiled down to a phone through volition elicitation and vGLP~\cite{shapiro2026volition}, Grassroots Logic Programs~\cite{shapiro2025glp}, their types~\cite{shapiro2026types} and their implementation~\cite{shapiro2026implementing}, and the Grassroots Super-App~\cite{shapiro2026gsg}, in which each platform is a mini-app.

\mypara{Person-to-person systems} Two deployed systems share the person-to-person architecture of grassroots platforms.  In Secure Scuttlebutt~\cite{tarr2019secure} each participant publishes to a single-writer, append-only log and replicates the logs of the identities they choose, so data flows along the edges of a social graph, as in grassroots social networks~\cite{shapiro2023gsn}.  Briar~\cite{briar2018} synchronises messages directly between its users' devices, with no central server, and over Bluetooth or Wi-Fi when the Internet is down.  Both are classified as grassroots by their essential agents, informally~\cite{shapiro2025characterising}.  Local-first software~\cite{kleppmann2019localfirst} keeps the primary copy of a person's data on their own devices, with servers holding secondary copies, and merges concurrent changes by conflict-free replicated data types that sync over any channel, a peer-to-peer connection included; in a grassroots platform, the copies of a person's state are held by the state custodians they designate among their friends~\cite{eitan2026securing}.

\section{Conclusion}
\label{sec:conclusion}

Grassroots platforms corral three AI risks of digital platforms.  A grassroots platform has no part whose capture is control of it, so to control it an AI must control the smartphones of its participants, one by one.  Under the contract of Section~\ref{sec:agents}, every AI agent has a principal and a developer, every act of an AI agent names its principal and requires its principal's attestation that it runs code signed by its developer, and its guardians and its developer are accountable for it.  Under the contract of Section~\ref{sec:provenance}, every item carries the signatures of its author and of every agent that forwarded it.

\bibliography{bib}

\appendix

\iffull
\section{The Framework, Recalled}\label{app:foundations}

This appendix recalls, in the words of~\cite{lewis2026volitional}, the mathematical foundations on which the paper is based; the proofs of the results recalled are in that paper.

Each agent consists of a \emph{person} and a \emph{machine} (e.g., smartphone) operated by the person.  A \emph{machine transaction} specifies what machines do; a \emph{volition guard} specifies which persons must be willing for the transaction to occur.

We assume a potentially infinite set of \emph{agents} $\Pi$, but consider only finite subsets of it, so we assume any particular set of agents $P \subset \Pi$ to be nonempty and finite.  We use $\subset$ to denote the strict subset relation and $\subseteq$ when equality is also possible.  As standard, we use $S^P$ to denote the set of all total functions from $P$ to $S$, and if $c\in S^P$ we use $c_p$ to denote the value of $c$ at $p\in P$.

\begin{definition}[Machine State, Configuration, Transaction, Volition-Guarded Multiagent Atomic Transaction~\cite{lewis2026volitional}]\label{def:mt}
Given an arbitrary set $S$ of \temph{machine states}, with a designated \temph{initial state} $s_0 \in S$, and agents $Q \subset \Pi$, a \temph{machine configuration} over $Q$ is a member of $S^Q$, and a \temph{machine transaction} over \temph{participants} $Q$ is a pair $c\rightarrow c' \in (S^Q)^2$ such that $c\ne c'$; it is \temph{unary} if $|Q|=1$.  Given such a machine transaction $t$, a \temph{volition-guarded multiagent atomic transaction} over $t$---henceforth, \temph{volition-guarded transaction}---is a pair $(t,Q')$ where $Q'\subseteq Q$ are its \temph{guards}.
\end{definition}

A volition-guarded transaction can be carried out only if its guards are willing; its participants include those whose state changes and those whose state is a precondition and does not change.

Distinct machine transactions can represent ``the same action'' in different configurations; this is captured with an equivalence relation on machine transactions.

\begin{definition}[Transaction Equivalence~\cite{lewis2026volitional}]\label{def:equivalence}
Given a set of machine transactions $R$, a \temph{transaction equivalence} is an equivalence relation $\sim$ on $R$ such that $t \sim t'$ implies $t$ and $t'$ have the same participants, and $R/{\sim}$ is countable.  We write $[t]$ for the equivalence class of $t$ under $\sim$.
\end{definition}

For example, all befriend$(p,q)$ transactions---differing only in the configurations in which they occur---form an equivalence class.  Next is the mathematical machinery allowing people to express their volitions regarding the classes of machine transactions their machines may participate in.

\begin{definition}[Agent State and Configuration~\cite{lewis2026volitional}]\label{def:agent-state}
Given agents $P$, states $S$ with initial state $s_0$, a set of machine transactions $T$ each over its own participants $Q\subseteq P$ and $S$, and equivalence $\sim$ on $T$, an \temph{agent state} is a pair $(V,m)\in \calA = (2^{T/\sim}  \times S)$ where  $V$ is its \temph{volitional state} and $m \in S$ its  \temph{machine state}.  The \temph{initial agent state} is $(\emptyset,s_0)$.  An \temph{agent configuration} $c$ over $P$, $S$, $T$, and $\sim$ is a member $c\in \calA^P$ in which $c^v_p \subseteq (T/{\sim})_p$ for every $p\in P$, where $(T/{\sim})_p$ denotes the classes in $T/{\sim}$ in which $p$ is a participant; we write $c^v_p$  for the volitional state and $c^m_p$ for the machine state of agent $p$ in $c$.
\end{definition}

\begin{definition}[Volitional Transaction~\cite{lewis2026volitional}]\label{def:vmat}
Given agents $P$, states $S$, machine transactions $T$ over $P$ and $S$, and equivalence $\sim$ on $T$:
\begin{enumerate}
    \item A \temph{volition change of agent $p\in P$} is a pair $c\rightarrow c'$ of agent configurations over $\{p\}$, $S$, $T$, and $\sim$ such that $c^v_p, c'{^v_p} \subseteq (T/{\sim})_p$ and $c^v_p \ne c'{^v_p}$, and $c^m_p = c'{^m_p}$.
    \item A \temph{volitional machine transaction} induced by a volition-guarded transaction $(t,Q')$, for some $t= (d\rightarrow d')\in T$ over $Q'\subseteq Q\subseteq P$, is a pair $c\rightarrow c'$ where $c\ne c'$ are agent configurations over $P$, $S$, $T$, and $\sim$ such that $[t]\in c^v_q$ for every $q\in Q'$; $c^m_p = d_p$ and $c'{^m_p} = d'_p$ for every $p\in Q$; $c^m_p = c'{^m_p}$ for every $p\in P\setminus Q$; and $c'{^v_p} = c^v_p \setminus \{[t]\}$ for every $p\in P$.
    \item A \temph{volitional transaction} is a volition change or a volitional machine transaction. 
\end{enumerate}
\end{definition}

When a volitional machine transaction induced by $(t,Q')$ is taken, the class $[t]$ is removed from every agent's volitional state: the will is fulfilled by any equivalent transaction.  A person may independently change their volitional state via volition changes, which may add or remove classes; beyond these, the framework removes a class from $c^v_p$ only upon fulfilment.

A person's not willing a transaction has, so far, no consequence: no volitional transaction is enabled by the absence of a will.  An implementation of volition-guarded transactions by communicating agents~\cite{shapiro2026volition} needs one, so that a machine asked for its person's will answers either way; it is the following, by which the machine answers for a person who does not will an act: in machine state $s$, if $p$ wills it $s'$ is taken, else $s''$.

\begin{definition}[Volition-Guarded If-Then-Else Transaction, Then-Branch, Else-Branch~\cite{lewis2026volitional}]\label{def:app-ite}
Given machine states $S$, a \temph{volition-guarded if-then-else transaction} over a single participant $p\in\Pi$ is a triple $s\rightarrow s'\,;\,s''$ \temph{guarded by} $p$, with $s,s',s''\in S$, $s\ne s'$ and $s\ne s''$; the unary machine transaction $s\rightarrow s'$ is its \temph{then-branch} and $s\rightarrow s''$ its \temph{else-branch}.
\end{definition}

\begin{definition}[Else Transaction~\cite{lewis2026volitional}]\label{def:app-else}
Given agents $P$, states $S$, machine transactions $T$ over $P$ and $S$, and equivalence $\sim$ on $T$, an \temph{else transaction} induced by a volition-guarded if-then-else transaction $s\rightarrow s'\,;\,s''$ guarded by $p\in P$, with $s\rightarrow s'\in T$, is a pair $c\rightarrow c'$ of agent configurations over $P$, $S$, $T$, and $\sim$ such that $[s\rightarrow s']\notin c^v_p$; $c^m_p = s$ and $c'{^m_p} = s''$; $c^m_q = c'{^m_q}$ for every $q\in P\setminus\{p\}$; and $c'{^v_q} = c^v_q$ for every $q\in P$.
\end{definition}

The then-branch, taken when $p$ wills it, is the unary volition-guarded transaction $(s\rightarrow s',\{p\})$ and induces volitional machine transactions per Definition~\ref{def:vmat}; a unary volition-guarded transaction is thus the special case with no else-branch.  The else transaction is taken when $p$ does not will the then-branch; it fulfils no will and removes no class.  The two are placed in one class below, so that liveness obliges the machine to answer either way; the obligation is the machine's, and it has no time bound.  Only a unary transaction has an else-branch: a binary one would let an agent that never wills it trigger it.

\mypara{Induced transition systems}  The following is a simplified variation, sufficient for the purpose of this work, on the foundations introduced in~\cite{shapiro2021multiagent}.

\begin{definition}[Transition System, Computation, Run, Safe, Live, Correct~\cite{lewis2026volitional}]\label{def:app-ts}
A \temph{transition system} is a tuple $TS=(S,s_0,T,{\sim})$, where: 
\begin{enumerate}
    \item $S$ is an arbitrary non-empty set, referred to as the set of \temph{states}.
    \item Some $s_0\in S$ is the designated \temph{initial state}. 
    \item $T\subseteq S^2$ is a set of \temph{correct transitions over} $S$, where each transition $t\in T$ is a pair $(s,s')$ of non-identical states $s\ne s'\in S$, also written as $t=s\rightarrow s'$.
    \item $\sim$ is a \temph{partial equivalence relation} on $T$: a symmetric and transitive relation on $T$, not necessarily reflexive.  Its domain $\{t\in T : t\sim t\}$ is partitioned into countably many \temph{liveness classes} $T/{\sim}$; a transition outside the domain belongs to no class.
\end{enumerate} 
A \temph{computation} of $TS$ is a (nonempty, potentially infinite) sequence of states $r= s_1,s_2,\cdots$; it is a \temph{run} of $TS$ if $s_1=s_0$.  A \temph{prefix} of a computation is a finite initial segment of it.
A computation $r= s_1,s_2,\ldots$ is \temph{safe}, also written $r\subseteq T$, if $s_i\rightarrow s_{i+1}\in T$ for every two consecutive states; and $s\xrightarrow{*}s'\subseteq T$ denotes the existence of a safe computation from $s$ to $s'$ (empty if $s=s'$).
A class $[t]\in T/{\sim}$ is \temph{enabled} in a state $s$ if $s\rightarrow s'\in[t]$ for some $s'\in S$.
A run $r$ is \temph{live} if no class $[t]\in T/{\sim}$ is enabled in every state of some suffix of $r$ with no member of $[t]$ occurring in the suffix.  A run is \temph{correct} if it is safe and live.
\end{definition}

A partial equivalence, rather than a total one, is used so that some transitions may carry no liveness requirement at all: only transitions in the domain of $\sim$ form classes and thereby incur a liveness obligation, while transitions outside the domain---belonging to no class---may occur in a correct run but are never required to.  A run with uncountably many classes enabled in every state of a suffix is not live.

\begin{lemma}[Extension~\cite{lewis2026volitional}]\label{lem:app-extension}
Every finite safe run of a transition system is a prefix of a correct run of it.
\end{lemma}

\begin{definition}[Multiagent Transition System~\cite{lewis2026volitional}]\label{def:app-mts}
Given agents $P \subset \Pi$ and an arbitrary set $S$ of \temph{states} with a designated \temph{initial state} $s_0\in S$, 
a \temph{multiagent transition system} over $P$ and $S$ is a transition system $TS= (C,c_0,T,{\sim})$ with \temph{configurations} $C:= S^P$, \temph{initial configuration}  $c_0:= \{s_0\}^P$, \temph{transitions} $T\subseteq C^2$ a set of transactions over $P$ and $S$, and $\sim$ a partial equivalence on $T$.
\end{definition}

Rather than specifying a multiagent transition system over a set of agents $P$ directly, it is specified via machine transactions (Definition~\ref{def:mt}).  A machine transaction over $Q\subseteq P$ defines a set of multiagent transitions over $P$ in which all members of $P\setminus Q$ are stationary:

\begin{definition}[Transaction Closure~\cite{lewis2026volitional}]\label{def:app-closure}
Let $P\subset \Pi$, $S$ a set of machine states, and $C:=S^P$.
For any transition or transaction $t = c\to c'$, we write $t_q := c_q\to c'_q$ and say $p$ is \temph{stationary} in $t$ if $c_p = c'_p$.
For a machine transaction $t=(c\rightarrow c')$ over $S$ with participants $Q$, the \temph{$P$-closure of $t$}, $t{\uparrow}P$, is the set of transitions over $P$ and $S$ defined by:
$$
t{\uparrow}P := \begin{cases} \{ t' \in C^2  : \forall q\in Q.(t_q = t'_q) \wedge {} & \text{if } Q\subseteq P \\
\qquad\forall p\in P\setminus Q.(p\text{ is stationary in }t')\} & \\
\emptyset & \text{otherwise}
\end{cases}
$$
If $R$ is a set of machine transactions, each $t\in R$ over some $Q$ and $S$, then the 
\temph{$P$-closure of $R$}, $R{\uparrow}P$, is the set of transitions over $P$ and $S$ defined by:
$$
R{\uparrow}P := \bigcup_{t\in R} t{\uparrow}P
$$
Given a relation ${\sim}$ on $R$, its \temph{$P$-closure} ${\sim}{\uparrow}P$ is the relation on $R{\uparrow}P$ with $\hat t \mathrel{({\sim}{\uparrow}P)} \hat t'$ iff $\hat t \in t{\uparrow}P$ and $\hat t' \in t'{\uparrow}P$ for some $t \sim t'$.
A transition over $P$ is \temph{unary} if it lies in the $P$-closure of a unary transaction.
\end{definition}

Namely, the closure over $P\supseteq Q$ of a machine transaction $t$ over $Q$ includes all transitions $t'$ over $P$ in which members of $Q$ do the same in $t$ and in $t'$, and the rest remain in their current (arbitrary) state.  The closure likewise carries any relation on transactions to one on the induced transitions.  If distinct transactions in $R$ have disjoint $P$-closures---as when every participant of every transaction in $R$ changes state---then each transition in $R{\uparrow}P$ has a unique inducing transaction, ${\sim}{\uparrow}P$ relates two transitions exactly when their inducing transactions are $\sim$-related, and ${\sim}{\uparrow}P$ is a partial equivalence whenever ${\sim}$ is.

A set $R$ of volition-guarded transactions with an equivalence $\sim$ on their underlying machine transactions induces a volitional multiagent transition system:

\begin{definition}[Volitional Multiagent Transition System~\cite{lewis2026volitional}]\label{def:app-vmts}
Given agents $P\subset \Pi$, machine states $S$ with initial state $s_0$, a set $R$ of volition-guarded transactions and volition-guarded if-then-else transactions---a volition-guarded if-then-else transaction $s\rightarrow s'\,;\,s''$ guarded by $p$ standing in $R$ for its then-branch $(s\rightarrow s',\{p\})$ together with its else-branch---such that every $(t,Q')\in R$ has the participants of $t$ contained in $P$, every guard $p$ of a volition-guarded if-then-else transaction is in $P$, and distinct underlying machine transactions, else-branches included, have disjoint $P$-closures, and an equivalence $\sim$ on the set $T_R := \{t : (t,Q')\in R\text{ for some }Q'\}$ of underlying machine transactions (a total equivalence, hence in particular a partial equivalence in the sense of Definition~\ref{def:app-ts}), such that no else-branch is in $T_R$ and volition-guarded if-then-else transactions with the same else-branch have $\sim$-equivalent then-branches, the \temph{volitional multiagent transition system induced by $(S,R,{\sim})$ over $P$} is the multiagent transition system $(\calA^P,c_0,T_V,{\sim_V})$ where:
\begin{enumerate}
    \item $\calA := 2^{T_R/\sim} \times S$ is the \temph{agent state space};
    \item $c_0 \in \calA^P$ is the \temph{initial agent configuration}, with $c_0{^v_p}=\emptyset$ and $c_0{^m_p}=s_0$ for every $p\in P$;
    \item $T_V$ consists of all transitions $e\to e'\in (\calA^P)^2$ of one of three forms: (\ia)~a \temph{volition change} of some $p\in P$---$e^v_p, e'{^v_p}\subseteq (T_R/{\sim})_p$ and $e^v_p \ne e'{^v_p}$, $e^m_p = e'{^m_p}$, and $e_r = e'_r$ for every $r\in P\setminus\{p\}$; (\ib)~a \temph{volitional machine transaction} induced by some volition-guarded transaction $(t,Q')\in R$ per Definition~\ref{def:vmat}(2); or (\ic)~an \temph{else transaction} induced by some volition-guarded if-then-else transaction in $R$ per Definition~\ref{def:app-else};
    \item $\sim_V$ is the restriction of ${\sim}{\uparrow}P$ (Definition~\ref{def:app-closure}) to the volitional machine transactions and else transactions, relating two of them whenever their inducing machine transactions are $\sim$-equivalent, where the inducing machine transaction of an else transaction is the then-branch $s\rightarrow s'$ of its volition-guarded if-then-else transaction, and leaves every volition-change transition outside its domain---so volition changes belong to no class and carry no liveness obligation.
\end{enumerate}
\end{definition}

The conditions on else-branches ensure that the machine transactions inducing a transition of $T_V$ are $\sim$-equivalent, so that $\sim_V$ is a partial equivalence.  The set $R$ of volition-guarded transactions that specifies a platform thus determines the volitional machine transitions of the induced volitional multiagent transition system, while volition-change transitions are freely available to every agent independent of $R$.

\begin{definition}[Enabled~\cite{lewis2026volitional}]\label{def:enabled}
Given a set of volition-guarded transactions, each $(t,Q')$ with $t = d \rightarrow d'$ a machine transaction over some  $Q' \subseteq Q\subseteq P$ and $S$, and an equivalence $\sim$ on machine transactions: the volition-guarded transaction $(t,Q')$ is \temph{machine-enabled} in agent configuration $c$ over $P$ if $c^m_p = d_p$ for every $p \in Q$, and \temph{enabled} if in addition $[t] \in c^v_q$ for every $q \in Q'$.  The else-branch of a volition-guarded if-then-else transaction $s\rightarrow s'\,;\,s''$ guarded by $p$ is \temph{enabled} in $c$ if $c^m_p = s$ and $[s\rightarrow s']\notin c^v_p$.  An equivalence class $[t]$ is \temph{enabled} in $c$ if some volition-guarded transaction $(t',Q')$ with $t' \in [t]$ is enabled in $c$, or the else-branch of some volition-guarded if-then-else transaction with then-branch $t'\in[t]$ is enabled in $c$.
\end{definition}

A volition-guarded transaction with an empty guard ($Q' = \emptyset$) requires no volitions and is enabled whenever its machine precondition is met.  The class of the then-branch of a volition-guarded if-then-else transaction guarded by $p$ is thus enabled whenever $p$'s machine is in its source state, whether or not $p$ wills it.  For the volitional multiagent transition system above, a class of volitional machine transactions and else transactions is enabled at a configuration in the sense of Definition~\ref{def:app-ts} exactly when some volition-guarded transaction or else-branch inducing it is enabled in the above sense; this determines which runs are live and correct per Definition~\ref{def:app-ts}.  Volition-change transitions belong to no class and so impose no liveness obligation; personal choices remain free.

\mypara{Protocols and grassroots protocols}  A protocol is a family of multiagent transition systems, one for each set of agents $P\subset \Pi$, which share an underlying set of local machine states $\calS$ with a designated initial state $s_0$.  A \emph{local-states function} maps every set of agents $P \subset \Pi$ to an arbitrary set of local machine states $S(P)\subset \calS$ that includes $s_0$ and satisfies $P\subset P' \subset \Pi \implies S(P) \subset S(P')$.

\begin{definition}[Protocol~\cite{lewis2026volitional}]\label{def:app-protocol}
A \temph{protocol} $\calF$ over a local-states function $S$ is a family of multiagent transition systems that has exactly one transition system $\calF(P) = (C(P),c_0(P),T(P),{\sim(P)})$ for every $P \subset \Pi$, with \temph{agent states} $\calA(P)$, configurations $C(P) := \calA(P)^P$, initial configuration $c_0(P)\in C(P)$, and partial equivalence $\sim(P)$ on $T(P)$ determined by the protocol, such that $P\subseteq P' \subset \Pi$ implies $\calA(P)\subseteq \calA(P')$ and $c_0(P)_p = c_0(P')_p$ for every $p\in P$.
\end{definition}

\begin{definition}[Interleaving~\cite{lewis2026volitional}]\label{def:app-interleaving}
Let $P, P' \subset \Pi$ be disjoint nonempty sets of agents, $r = c_0, c_1, \ldots$ a run of $\calF(P)$, and $r' = d_0, d_1, \ldots$ a run of $\calF(P')$.  An \temph{interleaving} of $r$ and $r'$ is a sequence $e_0, e_1, \ldots$ of configurations in $C(P \cup P')$ for which there exist non-decreasing sequences of indices $(i_k)_{k \geq 0}$ and $(j_k)_{k \geq 0}$ with $i_0 = j_0 = 0$ such that for every $k \geq 0$:
\begin{enumerate}
\item $(e_k)_p = (c_{i_k})_p$ for every $p \in P$,
\item $(e_k)_q = (d_{j_k})_q$ for every $q \in P'$,
\item if $e_{k+1}$ exists, then exactly one of:
  (\ia) $i_{k+1} = i_k + 1$ and $j_{k+1} = j_k$ (a $P$-step), or
  (\ib) $i_{k+1} = i_k$ and $j_{k+1} = j_k + 1$ (a $P'$-step).
\end{enumerate}
Moreover, if $r$ is finite of length $n$ then $i_k = n$ for some $k$, and if $r$ is infinite then for every $m \ge 0$ there is a $k$ with $i_k = m$; likewise for $r'$ and $(j_k)$.
\end{definition}

\begin{definition}[Interaction, Interactive Run, First Interaction~\cite{lewis2026volitional}]\label{def:app-interaction}
Let $Q\subset\Pi$.  For a configuration $c$ over $Q$ and $R\subseteq Q$, write $c|_R$ for the restriction of $c$ to $R$.  A transition $e\to e'$ of $\calF(Q)$ is an \temph{interaction between} $x\ne y\in Q$ if $e|_{\{x\}} \ne e'|_{\{x\}}$ and $e|_{\{y\}} \ne e'|_{\{y\}}$, and $e|_{\{x\}} \to e'|_{\{x\}}$ is not a transition of $\calF(\{x\})$ or $e|_{\{y\}} \to e'|_{\{y\}}$ is not a transition of $\calF(\{y\})$.  It is an \temph{interaction between} disjoint nonempty $P, P'\subseteq Q$ if it is an interaction between some $x\in P$ and some $y\in P'$.  A run $\hat r$ of $\calF(Q)$ is \temph{interactive between $P$ and $P'$} if some transition of $\hat r$ is an interaction between them; the \temph{first interaction} of $P$ and $P'$ in $\hat r$ is the first such transition.
\end{definition}

\begin{definition}[Oblivious, Interactive, Grassroots~\cite{lewis2026volitional}]\label{def:app-grassroots}
A  protocol $\calF$ is:
\begin{enumerate}
    \item \temph{oblivious} if for every disjoint nonempty $P, P' \subset \Pi$,
    every interleaving of a correct run of $\calF(P)$ and a correct run of $\calF(P')$ is a correct run of $\calF(P\cup P')$.
    \item  \temph{interactive} if for every $Q\subset\Pi$ and disjoint nonempty $P, P' \subseteq Q$, every prefix of a correct run of $\calF(Q)$ containing no interaction between $P$ and $P'$ is a prefix of a correct run of $\calF(Q)$ that contains one.
    \item \temph{grassroots} if it is oblivious and interactive.
\end{enumerate}
\end{definition}

Being oblivious means two disjoint groups coexist without interference: any interleaving of a correct run of one with a correct run of the other is a correct run of the combined system.  Being interactive means two groups that have not yet interacted can still do so: the combined system has a correct continuation in which a single transaction engages agents of both groups.  Agents that have taken part in a common interaction are joined; an instance is a maximal set of agents joined directly or indirectly, and two instances coalesce when they come to be one.

\begin{definition}[Interaction Graph, Instance, Coalescence~\cite{lewis2026volitional}]\label{def:app-coalescence}
Let $\calF$ be a protocol, $Q\subset\Pi$, and $r$ a prefix of a run of $\calF(Q)$.  The \temph{interaction graph} of $r$ has vertices $Q$ and an edge between $a\ne b\in Q$ whenever some transition of $r$ is an interaction between $a$ and $b$.  A nonempty $P\subseteq Q$ is an \temph{instance} in $r$ if it is a connected component of the interaction graph of $r$.  Disjoint nonempty $P, P'\subseteq Q$ \temph{coalesce} in a run $\hat r$ of $\calF(Q)$ if they are instances in some prefix of $\hat r$ and lie in one component of the interaction graph of a longer prefix of $\hat r$.
\end{definition}

\begin{definition}[Transactions Over a Local-State Function~\cite{lewis2026volitional}]\label{def:app-tblsf}
Let $S$ be a local-states function.
A set of transactions $R$ is \temph{over $S$} if every transaction $t\in R$
is a multiagent transition over $Q$ and $S(P')$ for some $Q \subseteq P'\subset \Pi$.  Given such a set $R$ and $P\subset \Pi$,
$
R(P) := \{ t\in R : t \text{ is over } Q \text{ and } S(P'), Q \subseteq P'\subseteq P\}
$.
\end{definition}

\begin{definition}[Volitional Transactions-Based Protocol~\cite{lewis2026volitional}]\label{def:app-vprotocol}
Let $S$ be a local-states function and $R$ a set of volition-guarded transactions over $S$ with equivalence $\sim$.
The \temph{protocol $\calF$ over $R$, $S$, and $\sim$} assigns to each set of agents $P\subset \Pi$ the volitional multiagent transition system $\calF(P)$ induced by $(S(P),R(P),{\sim})$ over $P$ (Definition~\ref{def:app-vmts}).  In particular, $C(P) = \calA(P)^P$ with agent state space $\calA(P) := 2^{T_{R(P)}/\sim} \times S(P)$---monotone in $P$, as Definition~\ref{def:app-protocol} requires---and $c_0(P)$ has $p$-component $(\emptyset,s_0)$ for every $p\in P$.
\end{definition}

The obliviousness results rest on the following invariant: in a group's own system an agent wills only transactions internal to the group, so the guard of a transaction that reaches another group cannot will it while the group runs alone, and the transaction is never enabled in an interleaving.

\begin{lemma}[Volitional Containment~\cite{lewis2026volitional}]\label{lem:app-containment}
Let $\calF$ be a volitional transactions-based protocol over a set of volition-guarded transactions $R$ with equivalence $\sim$.  For every $P\subset \Pi$, every configuration $c$ of $\calF(P)$, and every $p\in P$: $c^v_p \subseteq T_{R(P)}/\!\sim$, where $T_{R(P)} := \{t : (t,Q')\in R(P)\text{ for some }Q'\}$.
\end{lemma}

\begin{lemma}[Interleaving Safety~\cite{lewis2026volitional}]\label{lem:app-interleaving-safety}
Let $\calF$ be a volitional transactions-based protocol and $P, P'\subset \Pi$ disjoint and nonempty.  Every finite prefix of an interleaving of a correct run of $\calF(P)$ and a correct run of $\calF(P')$ is a finite safe run of $\calF(P\cup P')$.
\end{lemma}

\begin{proposition}[\cite{lewis2026volitional}]\label{prop:app-oblivious-criterion}
A volitional transactions-based protocol is oblivious provided that for every disjoint nonempty $P, P' \subset \Pi$, no equivalence class whose transactions have participants spanning both $P$ and $P'$ is ever enabled in any interleaving of correct runs of $\calF(P)$ and $\calF(P')$.
\end{proposition}

\begin{corollary}[Guarded Obliviousness~\cite{lewis2026volitional}]\label{cor:app-guarded}
A volitional transactions-based protocol over a set of volition-guarded transactions $R$ is oblivious if every volition-guarded transaction $(t,Q')\in R$ whose machine transaction $t$ has two or more participants has a nonempty guard, $Q'\ne\emptyset$.
\end{corollary}

In a run of $\calF(P\cup P')$ a member may will a transaction reaching the other group, which a group's own run forbids; the volitional notion constrains the resulting coupling:

\begin{definition}[Left~\cite{lewis2026volitional}]\label{def:app-left}
Let $\calF$ be a volitional transactions-based protocol over $R$, $Q\subset\Pi$, $c$ a configuration of $\calF(Q)$, and $x\in Q$.  Then $x$ \temph{has left} in $c$ if no $(t,Q')\in R(Q)$ with $x\in Q'$ is machine-enabled in any configuration $c'$ of $\calF(Q)$ with $c\xrightarrow{*}c'\subseteq T(Q)$.
\end{definition}

\begin{definition}[Interactive Among the Agents That Have Not Left~\cite{lewis2026volitional}]\label{def:app-left-interactive}
A volitional transactions-based protocol $\calF$ is \temph{interactive among the agents that have not left} if for every $Q\subset\Pi$ and disjoint nonempty $P, P'\subseteq Q$, every prefix of a correct run of $\calF(Q)$ containing no interaction between $P$ and $P'$, in whose last configuration some agent of $P$ and some agent of $P'$ have not left, is a prefix of a correct run of $\calF(Q)$ that contains one.
\end{definition}

An interactive protocol is interactive among the agents that have not left, the latter requiring a continuation of fewer prefixes.  A group all of whose agents have left can never be connected by an act one of them wills, and the definition asks nothing of it.

\begin{definition}[Volitionally Grassroots~\cite{lewis2026volitional}]\label{def:app-volitionally-grassroots}
A volitional transactions-based protocol $\calF$ is \temph{volitionally grassroots} if it is oblivious, interactive among the agents that have not left, and, for every disjoint nonempty $P, P'\subset\Pi$ and every safe run of $\calF(P\cup P')$ interactive between $P$ and $P'$, the first interaction of $P$ and $P'$ is induced by a volition-guarded transaction $(t,Q')$ with $Q'\cap P\ne\emptyset$ and $Q'\cap P'\ne\emptyset$.
\end{definition}

Two groups thus become connected only by an act a person of each wills, and a grassroots protocol meeting the condition on first interactions is volitionally grassroots.  An else transaction (Definition~\ref{def:app-else}) is over a single participant and so is never an interaction; a binary transaction with an else-branch would let an agent of another group, who never wills it, trigger it, and could be a first interaction willed by neither, which is why Definition~\ref{def:app-ite} admits only unary ones.

\fi

\section{The Speech-Act Schemas of the Contracts}
\label{app:transactions}

The speech-act schemas of Agents and Provenance Groups are in the repository, \path{programs/gai/Agents.glp} and \path{programs/gai/Provenance-Groups.glp} in the GLP repository, each the full schemas of its contract, compiled from its Logic pragmas and those of the contracts it is written from, Knowledge, Friendship, Guardians, Forwarding and Groups~\cite{shapiro2026formalising}, their records \path{cascade/Agents} and \path{cascade/Provenance-Groups}; the boxes below give them by the numbers of the clauses they realise, in the notation of Section~\ref{sec:agents-transactions}.  Agents writes $\mathit{ai\_agent}$ for the $\mathit{ward}$ of Guardians and $\mathit{principal}$ for its $\mathit{creator}$, and the guardians of an AI agent decide for it jointly.  A person holds $\mathit{person}$ from the act that establishes their identity and an AI agent $\mathit{ai\_agent}$ from its creation.  The schemas of Guardians:\ref{cl:creation} and Guardians:\ref{cl:joint} for a new guardian add the record of the attestation of the AI agent, the acts of an AI agent in Guardians:\ref{cl:ward-person} and Guardians:\ref{cl:ward-ward} require its attestation, and those for knowing add its record (Agents:\ref{cl:ai-acts}); the schemas of Forwarding:\ref{cl:item-creation} and Forwarding:\ref{cl:forwarding} are a person's, and an AI agent with a guardian signs and forwards by the two schemas of Agents:\ref{cl:ai-acts}; the developer signs the code of an AI agent and the principal the attestation by Forwarding:\ref{cl:item-creation} (Agents:\ref{cl:accountability} and Agents:\ref{cl:ai-acts}).  The schemas of Provenance Groups are those of Agents, the schema of Friendship:\ref{cl:unfriend} also clearing every listing and membership between the two friends (Groups:\ref{cl:joining}), with those of Groups, a message being an item its sender authored, and its own; every act of a group is a person's, and an AI agent takes none.  $\mathit{author}(x)$ is the first signer of $x$ and $\mathrm{sgn}(x,p)$ is $x$ signed over by $p$~\cite{shapiro2026formalising}; a group is named by its leader and a name, $\mathit{Grupo}$; and $\ast$ stands for every value.

\begin{contractbox}{Agents: Schemas of Knowledge}
\raggedright
\begin{enumerate}[label=\arabic*.]
\item[1.]
{$\mathit{identity}(\Alice)$, $\mathit{person}(\Alice)$ $\leftarrow$}
\sasbody{$\neg\mathit{identity}(\Alice)$ guarded~by~$\Alice$.}
\item[2.]
{$\mathit{knows}(\Alice, \Bob)$, $\mathit{knows}(\Bob, \Alice)$ $\leftarrow$}
\sasbody{$\mathit{person}(\Alice)$, $\mathit{person}(\Bob)$ guarded~by~$\Alice$, $\Bob$.}
\item[3.]
{$\mathit{knows}(\Bob, \Carol)$ $\leftarrow$}
\sasbody{$\mathit{knows}(\Alice, \Bob)$, $\mathit{knows}(\Bob, \Alice)$, $\mathit{knows}(\Alice, \Carol)$, $\mathit{person}(\Alice)$, $\mathit{person}(\Bob)$, $\mathit{person}(\Carol)$ guarded~by~$\Alice$, $\Bob$.}
\end{enumerate}
\end{contractbox}

\begin{contractbox}{Agents: Schemas of Friendship}
\raggedright
\begin{enumerate}[label=\arabic*.]
\item[1.]
{$\mathit{friend}(\Alice, \Bob)$, $\mathit{friend}(\Bob, \Alice)$ $\leftarrow$}
\sasbody{$\mathit{knows}(\Alice, \Bob)$, $\mathit{knows}(\Bob, \Alice)$, $\neg\mathit{friend}(\Alice, \Bob)$, $\neg\mathit{friend}(\Bob, \Alice)$, $\mathit{person}(\Alice)$, $\mathit{person}(\Bob)$ guarded~by~$\Alice$, $\Bob$.}
\item[2.]
{$\neg\mathit{friend}(\Alice, \Bob)$, $\neg\mathit{friend}(\Bob, \Alice)$ $\leftarrow$}
\sasbody{$\mathit{person}(\Alice)$, $\mathit{person}(\Bob)$, $\mathit{friend}(\Alice, \Bob)$, $\mathit{friend}(\Bob, \Alice)$ guarded~by~$\Alice$.}
\end{enumerate}
\end{contractbox}

\begin{contractbox}{Agents: Schemas of Guardians}
\raggedright
\begin{enumerate}[label=\arabic*.]
\item[1.]
{$\mathit{identity}(\Carol)$, $\mathit{ai\_agent}(\Carol)$, $\mathit{has\_guardian}(\Carol, \Alice)$, $\mathit{principal}(\Alice, \Carol)$, $\mathit{guardian}(\Alice, \Carol)$, $\mathit{knows}(\Alice, \Carol)$, $\mathit{knows}(\Carol, \Alice)$, $\mathit{friend}(\Alice, \Carol)$, $\mathit{friend}(\Carol, \Alice)$, $\mathit{attestation}(\Carol, \mathit{Proof})$, $\mathit{attestation\_record}(\Alice, \Carol, \mathit{Proof})$ $\leftarrow$}
\sasbody{$\mathit{person}(\Alice)$, $\neg\mathit{identity}(\Carol)$, $\mathit{item}(\Alice, \mathit{Proof})$, $\mathit{author}(\mathit{Proof}) = \Alice$ guarded~by~$\Alice$, $\Carol$.}
\item[2.]
{$\mathit{guardian}(\Bob, \Carol)$, $\mathit{has\_guardian}(\Carol, \Bob)$, $\mathit{knows}(\Bob, \Carol)$, $\mathit{knows}(\Carol, \Bob)$, $\mathit{friend}(\Bob, \Carol)$, $\mathit{friend}(\Carol, \Bob)$, $\mathit{attestation\_record}(\Bob, \Carol, \mathit{Proof})$ $\leftarrow$}
\sasbody{$\mathit{ai\_agent}(\Carol)$, $\neg\mathit{retired}(\Carol)$, $\mathit{person}(\Bob)$, $\mathit{has\_guardian}(\Carol, \Alice)$, $\mathit{friend}(\Bob, \Alice)$, $\neg\mathit{guardian}(\Bob, \Carol)$, $\neg\mathit{has\_guardian}(\Carol, \Bob)$, $\mathit{attestation}(\Carol, \mathit{Proof})$ guarded~by~the guardians of $\Carol$, $\Bob$.}
\smallskip
\sashead{$\neg\mathit{guardian}(\Bob, \Carol)$, $\neg\mathit{has\_guardian}(\Carol, \Bob)$ $\leftarrow$}
\sasbody{$\mathit{guardian}(\Bob, \Carol)$, $\mathit{has\_guardian}(\Carol, \Bob)$ guarded~by~$\Bob$.}
\smallskip
\sashead{$\neg\mathit{guardian}(\Bob, \Carol)$, $\neg\mathit{has\_guardian}(\Carol, \Bob)$ $\leftarrow$}
\sasbody{$\mathit{guardian}(\Bob, \Carol)$, $\mathit{has\_guardian}(\Carol, \Bob)$ guarded~by~the guardians of $\Carol$ other than $\Bob$.}
\smallskip
\sashead{$\neg\mathit{friend}(\Bob, \Carol)$, $\neg\mathit{friend}(\Carol, \Bob)$, $\neg\mathit{guardian}(\Bob, \Carol)$, $\neg\mathit{has\_guardian}(\Carol, \Bob)$ $\leftarrow$}
\sasbody{$\mathit{friend}(\Bob, \Carol)$, $\mathit{friend}(\Carol, \Bob)$, $\mathit{guardian}(\Bob, \Carol)$, $\mathit{has\_guardian}(\Carol, \Bob)$ guarded~by~$\Bob$.}
\item[3.]
{$\mathit{knows}(\Carol, \Bob)$, $\mathit{knows}(\Bob, \Carol)$, $\mathit{attestation\_record}(\Bob, \Carol, \mathit{Proof})$ $\leftarrow$}
\sasbody{$\mathit{ai\_agent}(\Carol)$, $\mathit{has\_guardian}(\Carol, \Alice)$, $\mathit{person}(\Bob)$, $\mathit{attestation}(\Carol, \mathit{Proof})$ guarded~by~$\Carol$, $\Bob$.}
\smallskip
\sashead{$\mathit{friend}(\Carol, \Bob)$, $\mathit{friend}(\Bob, \Carol)$ $\leftarrow$}
\sasbody{$\mathit{ai\_agent}(\Carol)$, $\neg\mathit{retired}(\Carol)$, $\mathit{has\_guardian}(\Carol, \Alice)$, $\mathit{person}(\Bob)$, $\mathit{knows}(\Carol, \Bob)$, $\mathit{knows}(\Bob, \Carol)$, $\mathit{friend}(\Alice, \Bob)$, $\mathit{friend}(\Bob, \Alice)$, $\neg\mathit{friend}(\Carol, \Bob)$, $\neg\mathit{friend}(\Bob, \Carol)$, $\mathit{attestation}(\Carol, \mathit{Proof})$ guarded~by~$\Carol$, $\Bob$, $\Alice$.}
\smallskip
\sashead{$\neg\mathit{friend}(\Carol, \Bob)$, $\neg\mathit{friend}(\Bob, \Carol)$ $\leftarrow$}
\sasbody{$\neg\mathit{has\_guardian}(\Carol, \Bob)$, $\mathit{friend}(\Carol, \Bob)$, $\mathit{friend}(\Bob, \Carol)$, $\mathit{attestation}(\Carol, \mathit{Proof})$ guarded~by~$\Carol$.}
\smallskip
\sashead{$\neg\mathit{friend}(\Bob, \Carol)$, $\neg\mathit{friend}(\Carol, \Bob)$ $\leftarrow$}
\sasbody{$\neg\mathit{guardian}(\Bob, \Carol)$, $\mathit{friend}(\Bob, \Carol)$, $\mathit{friend}(\Carol, \Bob)$ guarded~by~$\Bob$.}
\smallskip
\sashead{$\neg\mathit{friend}(\Carol, \Bob)$, $\neg\mathit{friend}(\Bob, \Carol)$ $\leftarrow$}
\sasbody{$\mathit{has\_guardian}(\Carol, \Alice)$, $\neg\mathit{guardian}(\Bob, \Carol)$, $\mathit{friend}(\Carol, \Bob)$, $\mathit{friend}(\Bob, \Carol)$ guarded~by~$\Alice$.}
\item[4.]
{$\mathit{knows}(\Carol, \Dave)$, $\mathit{knows}(\Dave, \Carol)$, $\mathit{attestation\_record}(\Carol, \Dave, \mathit{Seal})$, $\mathit{attestation\_record}(\Dave, \Carol, \mathit{Proof})$ $\leftarrow$}
\sasbody{$\mathit{ai\_agent}(\Carol)$, $\mathit{has\_guardian}(\Carol, \Alice)$, $\mathit{ai\_agent}(\Dave)$, $\mathit{has\_guardian}(\Dave, \Bob)$, $\mathit{attestation}(\Carol, \mathit{Proof})$, $\mathit{attestation}(\Dave, \mathit{Seal})$ guarded~by~$\Carol$, $\Dave$; $\Alice$ and $\Bob$ possibly one person.}
\smallskip
\sashead{$\mathit{friend}(\Carol, \Dave)$, $\mathit{friend}(\Dave, \Carol)$ $\leftarrow$}
\sasbody{$\mathit{ai\_agent}(\Carol)$, $\neg\mathit{retired}(\Carol)$, $\mathit{has\_guardian}(\Carol, \Alice)$, $\mathit{ai\_agent}(\Dave)$, $\neg\mathit{retired}(\Dave)$, $\mathit{has\_guardian}(\Dave, \Bob)$, $\mathit{knows}(\Carol, \Dave)$, $\mathit{knows}(\Dave, \Carol)$, $\mathit{knows}(\Alice, \Dave)$, $\mathit{friend}(\Alice, \Bob)$, $\mathit{friend}(\Bob, \Alice)$, $\neg\mathit{friend}(\Carol, \Dave)$, $\neg\mathit{friend}(\Dave, \Carol)$, $\mathit{attestation}(\Carol, \mathit{Proof})$, $\mathit{attestation}(\Dave, \mathit{Seal})$ guarded~by~$\Carol$, $\Dave$, $\Alice$, $\Bob$; $\Alice$ and $\Bob$ possibly one person.}
\smallskip
\sashead{$\mathit{friend}(\Carol, \Dave)$, $\mathit{friend}(\Dave, \Carol)$ $\leftarrow$}
\sasbody{$\mathit{ai\_agent}(\Carol)$, $\neg\mathit{retired}(\Carol)$, $\mathit{has\_guardian}(\Carol, \Alice)$, $\mathit{ai\_agent}(\Dave)$, $\neg\mathit{retired}(\Dave)$, $\mathit{has\_guardian}(\Dave, \Alice)$, $\mathit{knows}(\Carol, \Dave)$, $\mathit{knows}(\Dave, \Carol)$, $\neg\mathit{friend}(\Carol, \Dave)$, $\neg\mathit{friend}(\Dave, \Carol)$, $\mathit{attestation}(\Carol, \mathit{Proof})$, $\mathit{attestation}(\Dave, \mathit{Seal})$ guarded~by~$\Carol$, $\Dave$, $\Alice$.}
\item[5.]
{$\mathit{retired}(\Carol)$, $\neg\mathit{guardian}(\ast, \Carol)$, $\neg\mathit{has\_guardian}(\Carol, \ast)$, $\neg\mathit{friend}(\Carol, \ast)$, $\neg\mathit{friend}(\ast, \Carol)$ $\leftarrow$}
\sasbody{$\neg\mathit{retired}(\Carol)$ guarded~by~the guardians of $\Carol$.}
\end{enumerate}
\end{contractbox}

\begin{contractbox}{Agents: Schemas of Forwarding}
\raggedright
\begin{enumerate}[label=\arabic*.]
\item[1.]
{$\mathit{item}(\Alice, \mathit{Item})$ $\leftarrow$}
\sasbody{$\neg\mathit{item}(\Alice, \mathit{Item})$, $\mathit{person}(\Alice)$ guarded~by~$\Alice$.}
\item[2.]
{$\mathit{item}(\Bob, \mathrm{sgn}(\mathit{Item},\Alice))$ $\leftarrow$}
\sasbody{$\mathit{item}(\Alice, \mathit{Item})$, $\mathit{friend}(\Alice, \Bob)$, $\mathit{friend}(\Bob, \Alice)$, $\mathit{person}(\Alice)$ guarded~by~$\Alice$.}
\end{enumerate}
\end{contractbox}

\begin{contractbox}{Agents: Schemas of Agents}
\raggedright
\begin{enumerate}[label=\arabic*.]
\item[2.]
{$\mathit{item}(\mathit{Alf}, \mathit{Item})$ $\leftarrow$}
\sasbody{$\mathit{ai\_agent}(\mathit{Alf})$, $\mathit{attestation}(\mathit{Alf}, \mathit{Proof})$, $\mathit{has\_guardian}(\mathit{Alf}, \Alice)$, $\neg\mathit{item}(\mathit{Alf}, \mathit{Item})$ guarded~by~$\mathit{Alf}$.}
\smallskip
\sashead{$\mathit{item}(\Bob, \mathrm{sgn}(\mathit{Item},\mathit{Alf}))$ $\leftarrow$}
\sasbody{$\mathit{ai\_agent}(\mathit{Alf})$, $\mathit{attestation}(\mathit{Alf}, \mathit{Proof})$, $\mathit{has\_guardian}(\mathit{Alf}, \Alice)$, $\mathit{item}(\mathit{Alf}, \mathit{Item})$, $\mathit{friend}(\mathit{Alf}, \Bob)$, $\mathit{friend}(\Bob, \mathit{Alf})$ guarded~by~$\mathit{Alf}$.}
\end{enumerate}
\end{contractbox}

\begin{contractbox}{Provenance Groups: Schemas of Groups}
\raggedright
\begin{enumerate}[label=\arabic*.]
\item[2.]
{$\neg\mathit{friend}(\Alice, \Bob)$, $\neg\mathit{friend}(\Bob, \Alice)$, $\neg\mathit{member}(\Alice, \ast, \Bob)$, $\neg\mathit{member}(\Bob, \ast, \Alice)$, $\neg\mathit{listed}(\Alice, \Bob, \ast)$, $\neg\mathit{listed}(\Bob, \Alice, \ast)$ $\leftarrow$}
\sasbody{$\mathit{person}(\Alice)$, $\mathit{person}(\Bob)$, $\mathit{friend}(\Alice, \Bob)$, $\mathit{friend}(\Bob, \Alice)$ guarded~by~$\Alice$.}
\item[1.]
{$\mathit{item}(\Bob, \mathit{Text})$ $\leftarrow$}
\sasbody{$\mathit{friend}(\Alice, \Bob)$, $\mathit{friend}(\Bob, \Alice)$, $\mathit{person}(\Alice)$, $\mathit{item}(\Alice, \mathit{Text})$, $\mathit{author}(\mathit{Text}) = \Alice$ guarded~by~$\Alice$; $\Alice$ and $\Bob$ possibly one person.}
\item[2.]
{$\mathit{leader}(\Alice, \mathit{Grupo})$, $\mathit{member}(\Alice, \mathit{Grupo}, \Alice)$, $\mathit{listed}(\Alice, \Alice, \mathit{Grupo})$ $\leftarrow$}
\sasbody{$\neg\mathit{leader}(\Alice, \mathit{Grupo})$, $\mathit{person}(\Alice)$ guarded~by~$\Alice$.}
\smallskip
\sashead{$\neg\mathit{leader}(\Alice, \mathit{Grupo})$, $\neg\mathit{listed}(\Alice, \ast, \mathit{Grupo})$, $\neg\mathit{member}(\ast, \mathit{Grupo}, \Alice)$ $\leftarrow$}
\sasbody{$\mathit{leader}(\Alice, \mathit{Grupo})$, $\mathit{person}(\Alice)$ guarded~by~$\Alice$.}
\item[3.]
{$\mathit{member}(\Bob, \mathit{Grupo}, \Alice)$, $\mathit{listed}(\Alice, \Bob, \mathit{Grupo})$ $\leftarrow$}
\sasbody{$\mathit{leader}(\Alice, \mathit{Grupo})$, $\mathit{friend}(\Alice, \Bob)$, $\mathit{friend}(\Bob, \Alice)$, $\neg\mathit{member}(\Bob, \mathit{Grupo}, \ast)$, $\mathit{person}(\Alice)$, $\mathit{person}(\Bob)$ guarded~by~$\Alice$, $\Bob$; $\Alice$ and $\Bob$ possibly one person.}
\smallskip
\sashead{$\neg\mathit{member}(\Bob, \mathit{Grupo}, \Alice)$, $\neg\mathit{listed}(\Alice, \Bob, \mathit{Grupo})$ $\leftarrow$}
\sasbody{$\mathit{leader}(\Alice, \mathit{Grupo})$, $\mathit{member}(\Bob, \mathit{Grupo}, \Alice)$, $\mathit{listed}(\Alice, \Bob, \mathit{Grupo})$, $\mathit{person}(\Alice)$ guarded~by~$\Alice$; $\Alice$ and $\Bob$ possibly one person.}
\smallskip
\sashead{$\neg\mathit{member}(\Bob, \mathit{Grupo}, \Alice)$, $\neg\mathit{listed}(\Alice, \Bob, \mathit{Grupo})$ $\leftarrow$}
\sasbody{$\mathit{member}(\Bob, \mathit{Grupo}, \Alice)$, $\mathit{listed}(\Alice, \Bob, \mathit{Grupo})$, $\mathit{person}(\Bob)$ guarded~by~$\Bob$; $\Alice$ and $\Bob$ possibly one person.}
\item[4.]
{$\mathit{posted}(\Alice, \mathit{Text}, \Bob, \mathit{Grupo})$ $\leftarrow$}
\sasbody{$\mathit{member}(\Bob, \mathit{Grupo}, \Alice)$, $\neg\mathit{posted}(\Alice, \mathit{Text}, \Bob, \mathit{Grupo})$, $\mathit{person}(\Bob)$, $\mathit{item}(\Bob, \mathit{Text})$, $\mathit{author}(\mathit{Text}) = \Bob$ guarded~by~$\Bob$; $\Alice$ and $\Bob$ possibly one person.}
\smallskip
\sashead{$\mathit{posted}(\Alice, \mathit{Text}, \Alice, \mathit{Grupo})$ $\leftarrow$}
\sasbody{$\mathit{member}(\Alice, \mathit{Grupo}, \Alice)$, $\neg\mathit{posted}(\Alice, \mathit{Text}, \Alice, \mathit{Grupo})$, $\mathit{person}(\Alice)$, $\mathit{item}(\Alice, \mathit{Text})$, $\mathit{author}(\mathit{Text}) = \Alice$ guarded~by~$\Alice$.}
\smallskip
\sashead{$\mathit{group\_delivered}(\Alice, \Bob, \mathit{Text}, \mathit{Grupo}, \Carol)$, $\mathit{item}(\Carol, \mathrm{sgn}(\mathit{Text},\Alice))$ $\leftarrow$}
\sasbody{$\mathit{leader}(\Alice, \mathit{Grupo})$, $\mathit{posted}(\Alice, \mathit{Text}, \Bob, \mathit{Grupo})$, $\mathit{listed}(\Alice, \Carol, \mathit{Grupo})$, $\neg\mathit{group\_delivered}(\Alice, \Bob, \mathit{Text}, \mathit{Grupo}, \Carol)$, $\mathit{author}(\mathit{Text}) \neq \Alice$ guarded~by~no~one; $\Alice$ and $\Bob$, $\Alice$ and $\Carol$, $\Bob$ and $\Carol$ possibly one person.}
\smallskip
\sashead{$\mathit{group\_delivered}(\Alice, \Bob, \mathit{Text}, \mathit{Grupo}, \Alice)$, $\mathit{item}(\Alice, \mathrm{sgn}(\mathit{Text},\Alice))$ $\leftarrow$}
\sasbody{$\mathit{leader}(\Alice, \mathit{Grupo})$, $\mathit{posted}(\Alice, \mathit{Text}, \Bob, \mathit{Grupo})$, $\neg\mathit{group\_delivered}(\Alice, \Bob, \mathit{Text}, \mathit{Grupo}, \Alice)$, $\mathit{author}(\mathit{Text}) \neq \Alice$ guarded~by~no~one; $\Alice$ and $\Bob$ possibly one person.}
\end{enumerate}
\end{contractbox}

\begin{contractbox}{Provenance Groups: Schemas of Provenance Groups}
\raggedright
\begin{enumerate}[label=\arabic*.]
\item[1.]
{$\mathit{item}(\Bob, \mathit{Text})$ $\leftarrow$}
\sasbody{$\mathit{ai\_agent}(\mathit{Alf})$, $\mathit{attestation}(\mathit{Alf}, \mathit{Proof})$, $\mathit{has\_guardian}(\mathit{Alf}, \Alice)$, $\mathit{item}(\mathit{Alf}, \mathit{Text})$, $\mathit{friend}(\mathit{Alf}, \Bob)$, $\mathit{friend}(\Bob, \mathit{Alf})$ guarded~by~$\mathit{Alf}$.}
\smallskip
\sashead{$\mathit{group\_delivered}(\Alice, \Alice, \mathit{Text}, \mathit{Grupo}, \Carol)$, $\mathit{item}(\Carol, \mathit{Text})$ $\leftarrow$}
\sasbody{$\mathit{leader}(\Alice, \mathit{Grupo})$, $\mathit{posted}(\Alice, \mathit{Text}, \Alice, \mathit{Grupo})$, $\mathit{listed}(\Alice, \Carol, \mathit{Grupo})$, $\mathit{author}(\mathit{Text}) = \Alice$, $\neg\mathit{group\_delivered}(\Alice, \Alice, \mathit{Text}, \mathit{Grupo}, \Carol)$ guarded~by~no~one.}
\item[2.]
{$\mathit{posted}(\Alice, \mathrm{sgn}(\mathit{Item},\Bob), \Bob, \mathit{Grupo})$ $\leftarrow$}
\sasbody{$\mathit{person}(\Bob)$, $\mathit{member}(\Bob, \mathit{Grupo}, \Alice)$, $\neg\mathit{leader}(\Bob, \mathit{Grupo})$, $\mathit{item}(\Bob, \mathit{Item})$ guarded~by~$\Bob$.}
\end{enumerate}
\end{contractbox}

{\sloppy We name the acts as the repository does, and the boxes list the schemas of a clause in this order: Establish, Acquaint and Introduce (Knowledge:\ref{cl:identity} to Knowledge:\ref{cl:introduction}); Befriend and Person unfriend (Friendship:\ref{cl:friendship} and Friendship:\ref{cl:unfriend}); Enter (Guardians:\ref{cl:creation}); Make guardian, Resign, End guardian and Guardian unfriend (Guardians:\ref{cl:joint}); AI agent acquaint, AI agent befriend, AI agent end, Unfriend and AI agent unfriend (Guardians:\ref{cl:ward-person}); AI agents acquaint, AI agents befriend and AI agents of one guardian befriend (Guardians:\ref{cl:ward-ward}); Retire (Guardians:\ref{cl:end-supervision}); Sign and Forward (Forwarding:\ref{cl:item-creation} and Forwarding:\ref{cl:forwarding}); Agent sign and Agent forward (Agents:\ref{cl:ai-acts}); Person unfriend, with Groups:\ref{cl:joining}, Message (Groups:\ref{cl:message}), Create and Close (Groups:\ref{cl:group}), Join, Remove and Leave (Groups:\ref{cl:joining}), and Post, Post own, Deliver and Deliver leader (Groups:\ref{cl:group-message}); and Agent message and Deliver own (Provenance Groups:\ref{cl:agent-message}) and Post forward (Provenance Groups:\ref{cl:post-forward}).\par}

The guardians of an AI agent decide for it jointly: an act guarded by the guardians of Carol is guarded by every $g$ with $\mathit{has\_guardian}(\Carol,g)$, each holding $\mathit{guardian}(g,\Carol)$ and a record of Carol's attestation~\cite{shapiro2026formalising,shapiro2026volition}.  The schema of Guardians:\ref{cl:end-supervision} clears, by $\ast$, every guardianship and friendship of the AI agent, at it and at each of its guardians and friends.  Alice, the guardian an AI agent holds recorded in its acts of knowing, signing, forwarding and messaging, is an agent variable there and takes no part in the act.  The acts of Knowledge:\ref{cl:meeting}, Guardians:\ref{cl:ward-person} and Guardians:\ref{cl:ward-ward} for knowing neither require nor forbid that the two know each other.

\begin{table*}[t]
\caption{The predicates of the contract; the first argument of each is its holder $p$.}
\label{tab:local-state}
\footnotesize
\begin{tabular}{@{}p{0.3\linewidth}p{0.66\linewidth}@{}}
\hline
predicate & meaning \\
\hline
$\mathit{identity}(p)$ & $p$ has a genuine digital identity \\
$\mathit{person}(p)$ & $p$ is a person \\
$\mathit{ai\_agent}(p)$ & $p$ is an AI agent \\
$\mathit{knows}(p,q)$ & $p$ knows $q$, by $q$'s public key \\
$\mathit{friend}(p,q)$ & $q$ is a friend of $p$ \\
$\mathit{principal}(p,q)$ & $p$ is the principal of the AI agent $q$ \\
$\mathit{guardian}(p,q)$, $\mathit{has\_guardian}(p,q)$ & $p$ is a guardian of the AI agent $q$; $q$ is a guardian of the AI agent $p$ \\
$\mathit{attestation}(p,y)$ & $y$ is the attestation of the AI agent $p$ \\
$\mathit{attestation\_record}(p,q,y)$ & $p$ holds a record of the attestation $y$ of the AI agent $q$ \\
$\mathit{retired}(p)$ & the AI agent $p$ was retired \\
$\mathit{item}(p,x)$ & $p$ holds the item $x$, signed by its author and by every agent that forwarded it, in order \\
\hline
\end{tabular}
\end{table*}
\iffull
\begin{table*}[t]
\caption{The atoms the acts of the example add and delete.}
\label{tab:example}
\footnotesize
\begin{tabular}{p{0.2\linewidth}p{0.26\linewidth}p{0.18\linewidth}p{0.28\linewidth}}
Acts & Alice & Bob & Alf \\
\hline
Knowledge:\ref{cl:identity}, by Alice and by Bob & $\mathit{identity}$, $\mathit{person}$ & $\mathit{identity}$, $\mathit{person}$ & \\
Forwarding:\ref{cl:item-creation}, Alice signing $y$ & $\mathit{item}(y)$ & & \\
Guardians:\ref{cl:creation} & $\mathit{principal}(\mathit{Alf})$, $\mathit{guardian}(\mathit{Alf})$, $\mathit{knows}(\mathit{Alf})$, $\mathit{friend}(\mathit{Alf})$, $\mathit{attestation\_record}(\mathit{Alf},y)$ & & $\mathit{identity}$, $\mathit{ai\_agent}$, $\mathit{has\_guardian}(\Alice)$, $\mathit{knows}(\Alice)$, $\mathit{friend}(\Alice)$, $\mathit{attestation}(y)$ \\
Knowledge:\ref{cl:meeting} & $\mathit{knows}(\Bob)$ & $\mathit{knows}(\Alice)$ & \\
Guardians:\ref{cl:ward-person}, knowing & & $\mathit{knows}(\mathit{Alf})$, $\mathit{attestation\_record}(\mathit{Alf},y)$ & $\mathit{knows}(\Bob)$ \\
Friendship:\ref{cl:friendship} & $\mathit{friend}(\Bob)$ & $\mathit{friend}(\Alice)$ & \\
Guardians:\ref{cl:ward-person}, a relation & & $\mathit{friend}(\mathit{Alf})$ & $\mathit{friend}(\Bob)$ \\
Agents:\ref{cl:ai-acts} and Provenance Groups:\ref{cl:agent-message}, Alf signing and sending $x$ & & $\mathit{item}(x)$ & $\mathit{item}(x)$ \\
Guardians:\ref{cl:end-supervision} & $\neg\mathit{guardian}(\mathit{Alf})$, $\neg\mathit{friend}(\mathit{Alf})$ & $\neg\mathit{friend}(\mathit{Alf})$ & $\mathit{retired}$, $\neg\mathit{friend}(\ast)$, $\neg\mathit{has\_guardian}(\ast)$ \\
\end{tabular}
\end{table*}
\fi
\mypara{Transaction equivalence}  The equivalence~$\sim$ is that of~\cite{shapiro2026formalising}: one schema at one admissible binding determines one class, whose members differ only in the configuration in which they occur.

\section{The Contracts Certified}
\label{app:proofs}

Contracts with wards, agents who act only while they have a guardian, are defined in~\cite{shapiro2026formalising}, with conditions on the schemas of a contract, decided by the checker in the repository \repo: a contract has \emph{kinds}, a mark for a person and a mark for each kind of ward, held from the act that establishes an agent's identity; \emph{guardianships}, the two records of a guardian, at the guardian and at the ward, with every guarding role a person's, a ward's who has a guardian, or that of an agent who \emph{can act}, a person or a ward with a guardian; an introductory act \emph{unobstructed among the free}, by which two persons come to know each other; and \emph{relay acts}, by which a ward with a guardian and another agent come to know each other, the ward presenting what it carries and the other recording it.  A contract with these has an \emph{introduction with wards}, and the protocol realising a contract with an introduction with wards, traceable provenance and volition is volitionally grassroots (Theorem Grassroots realisation with wards of~\cite{shapiro2026formalising}).  Guardians is certified so there, and Agents and Provenance Groups are certified here, on the schemas of Appendix~\ref{app:transactions}.\full{}{  The proofs are in the full version of this paper.}

\begin{proposition}[Grassroots]\label{prop:grassroots}
The protocol realising Agents is volitionally grassroots.
\end{proposition}

\iffull
\begin{proof}\sloppy
The checker certifies the schemas of Agents, \path{programs/gai/Agents.glp}, as syntactically grassroots with wards, its verdict \path{cascade/Agents/verdict.txt}.  \emph{Kinds.}  The kinds are $\mathit{identity}$, $\mathit{person}$ and $\{\mathit{ai\_agent}\}$: no schema deletes any of the three, Establish adds $\mathit{identity}$ with $\mathit{person}$ and Enter adds it at the AI agent with $\mathit{ai\_agent}$, each forbidding it there, and no other schema adds one of them.  \emph{Guardianships.}  The guardianship of $\mathit{ai\_agent}$ has $\mathit{guardian}$ and $\mathit{has\_guardian}$ for its two records, which Enter and Make guardian add together and Resign, End guardian, Guardian unfriend and Retire delete together; every guarding role forbids $\mathit{identity}$, requires $\mathit{person}$, or holds in its body that its agent can act, an AI agent with a guardian.  \emph{Introduction.}  Acquaint is unobstructed among the free, as in Guardians.  AI agent acquaint is a relay act with the ward role Carol and the free role Bob, and AI agents acquaint one with the ward roles Carol and Dave: a ward role requires $\mathit{ai\_agent}$, $\mathit{has\_guardian}$ of an agent variable and $\mathit{attestation}$ of a speech-act variable, which is what a ward carries, since Enter, the one schema adding $\mathit{ai\_agent}$, adds $\mathit{attestation}(\mathit{Proof})$ at the AI agent, and no schema deletes an atom of $\mathit{attestation}$; the free role requires $\mathit{person}$; and AI agent acquaint adds the record $\mathit{attestation\_record}$ at its free role, and AI agents acquaint the record of each ward role at the other.  \emph{Traceable provenance and volition.}  The checker decides the two conditions on the schemas, and its verdict has both.  So Agents has an introduction with wards, traceable provenance and volition, and the protocol realising it is volitionally grassroots by Theorem Grassroots realisation with wards of~\cite{shapiro2026formalising}.
\end{proof}
\fi

\begin{proposition}[The social network with provenance is grassroots]\label{prop:gsnp-grassroots}
The protocol realising Provenance Groups is volitionally grassroots.
\end{proposition}

\iffull
\begin{proof}\sloppy
The checker certifies the schemas of Provenance Groups, \path{programs/gai/Provenance-Groups.glp}, as syntactically grassroots with wards, its verdict \path{cascade/Provenance-Groups/verdict.txt}, with the kinds, the guardianship, the introductory act and the relay acts of Agents, as in the proof of Proposition~\ref{prop:grassroots}.  Provenance Groups has the schemas of Agents, Person unfriend also clearing every listing and membership between the two, with Message, Agent message and the schemas of a group, which name a group by its leader and a name~\cite{shapiro2026formalising}.  The added schemas add and delete none of $\mathit{identity}$, $\mathit{person}$, $\mathit{ai\_agent}$, $\mathit{guardian}$, $\mathit{has\_guardian}$, $\mathit{attestation}$ and $\mathit{attestation\_record}$; the role of Agent message holds in its body that its agent, an AI agent, has a guardian; every guarding role of Message and of a schema of a group requires $\mathit{person}$; and Deliver, Deliver leader and Deliver own have no guard.  So Provenance Groups has an introduction with wards, traceable provenance and volition, and the protocol realising it is volitionally grassroots by Theorem Grassroots realisation with wards of~\cite{shapiro2026formalising}.
\end{proof}
\fi

\end{document}